\documentclass{article}

\usepackage[preprint]{neurips_2026}

\usepackage[utf8]{inputenc}
\usepackage[T1]{fontenc}
\usepackage[hidelinks,hypertexnames=false]{hyperref}
\usepackage{url}
\usepackage{microtype}
\usepackage{graphicx}
\usepackage{booktabs}
\usepackage{amsmath}
\usepackage{amssymb}
\usepackage{mathtools}
\usepackage{xcolor}
\usepackage{array}
\usepackage{float}
\usepackage{capt-of}

\newcommand{\includegraphicsmaybe}[2][]{%
  \IfFileExists{#2.pdf}{\includegraphics[#1]{#2}}{%
  \IfFileExists{#2.png}{\includegraphics[#1]{#2}}{%
  \fbox{\begin{minipage}{0.88\linewidth}\centering Missing figure: \texttt{\detokenize{#2}}\end{minipage}}}}%
}

\newcommand{\KL}{K_L}
\newcommand{\KH}{K_H}
\newcommand{\zH}{z_H}
\newcommand{\zL}{z_L}

\definecolor{geompath}{HTML}{DFF2FF}
\definecolor{geomwall}{HTML}{222222}
\definecolor{geombox}{HTML}{FFF2CC}
\definecolor{geomrow}{HTML}{E9F5FF}
\definecolor{geomcol}{HTML}{F4E9FF}
\definecolor{geomarcA}{HTML}{8DD3C7}
\definecolor{geomarcB}{HTML}{FB8072}
\definecolor{geomarcC}{HTML}{BEBADA}

\usepackage[capitalize,noabbrev]{cleveref}

\title{Interaction Locality in Hierarchical Recursive Reasoning}

\author{%
  Yosuke Miyanishi, Tetsuro Morimura \\
  CyberAgent Inc. \\
  \texttt{miyanishi\_yosuke@cyberagent.co.jp}
}

\begin{document}
\maketitle

\begin{abstract}
Spatial reasoning requires both location-bound computation and location-invariant structure: agents must make local moves while preserving route, object, or constraint-level plans. We propose interaction locality, a task-geometry-aware framework for measuring whether information flow stays within nearby cells or semantic segments, or crosses them. We instantiate the framework with sparse-autoencoder feature ablations and finite-noise activation patching, with structural Jacobian and attention checks reported in the appendix, and apply it to HRM and TRM, two compact hierarchical and recursive reasoning models, on Maze-Hard, Sudoku Extreme, and ARC-AGI. Across these models, activation patching gives the clearest architectural fingerprint: high-level recurrent states tend to write information within nearby cells or same-segment units, while repeated recursive updates accumulate these local writes into broader solution structure. This pattern holds across maze paths, Sudoku constraints, and ARC-AGI object neighborhoods, with the strongest concentration in TRM. To test whether interaction locality extends beyond toy-yet-challenging grid benchmarks, we also apply it to MTU3D, a large-scale embodied 3D scene-grounding model. In this MTU3D setting, causal spatial locality appears primarily at the transition where visual scene features are handed to the downstream grounding module, rather than uniformly throughout the visual encoder. This contrast suggests that the local-to-global handoff observed in HRM and TRM is tied to explicit recursive reasoning dynamics, while embodied 3D models may concentrate causal spatial structure at module boundaries. Interaction locality turns the intuitive local-execution/global-planning story into a reproducible measurement framework for recursive and embodied spatial reasoning.
\end{abstract}

\section{Introduction}
\label{sec:intro}

Spatial reasoning is central to real-world agents: robots manipulate objects,
vehicles plan around obstacles, and navigation systems search maps while
maintaining route-level intent. These tasks require a characteristic separation
of scale. A model must preserve location-specific facts such as walls, cells,
colors, and immediate moves, while also forming location-invariant structure such
as corridors, constraint houses, objects, and global plans. Because spatial
reasoning systems can affect physical actions, interpretability should explain
how local and global information flow through the computation.

Compact recursive architectures provide a focused testbed for this question.
The Hierarchical Reasoning Model (HRM; \cite{hrm2024}) and Tiny Recursive Model
(TRM; \cite{trm2024}) solve challenging Maze-Hard, Sudoku Extreme, and ARC-AGI
tasks by repeatedly applying small Transformer modules \citep{vaswani2017attention}.
HRM has separate high-level and low-level modules, whereas TRM reuses a single
module across recursive calls. It is tempting to equate the H state with global
planning and the L state with local refinement, but this is a hypothesis, not a 
mechanistic measurement: a label does not determine the spatial reach of an update.

We make this hypothesis testable through \emph{interaction locality}. For each
task, the external geometry defines both fine sites and coarser semantic
segments: maze cells and corridors, Sudoku cells and houses, and ARC-AGI
foreground cells and objects. We ask whether hidden features, update kernels, or
finite perturbations stay within these neighborhoods or cross them. This gives a
single coordinate system for comparing local movement, constraint propagation,
object-level aggregation, and route-level planning across architectures.

Our contribution is a unified locality framework and a controlled analysis of
HRM/TRM under a layered probe suite. Sparse autoencoder (SAE) feature ablations
expose human-readable segment effects, finite-noise activation patching provides
the main causal test of spatial reach, and appendix structural Jacobian/attention
checks report the corresponding linearized or architectural topology. The probes agree on a sharper
conclusion than the informal \textit{H is global} story: across Maze, Sudoku, and
ARC-AGI, H-level writes are often more spatially concentrated than L-level
writes, but cross-cycle propagation can still carry those summaries broadly
through the recursive state. We also stress-test the framework beyond toy grids
by applying it to MTU3D, a 3D embodied navigation and grounding model
\citep{mtu3d2025}, on ScanNet indoor scenes \citep{dai2017scannet}. There,
causal spatial locality appears at the visual-to-grounding handoff but largely
disappears inside the unified encoder.

\section{Background and related work}
\label{sec:background}

Both HRM and TRM maintain carry states $(\zH,\zL)\in\mathbb{R}^{B\times T\times D}$
through repeated reasoning steps. We write cycle labels such as H1L1 as
zero-indexed high-phase/low-call locations. At a low call,
\begin{equation}
  \zL^{(n+1)}=f_L(\zL^{(n)},\zH^{(n)}+\mathbf e),\qquad
  \zH^{(n+1)}=f_H(\zH^{(n)},\zL^{(n+1)}),
  \label{eq:zl}
\end{equation}
where $\mathbf e$ is the token embedding stream. HRM uses distinct $f_L$ and
$f_H$ modules; TRM shares a single module across H/L calls, so any H/L
distinction must arise from call context and state trajectory rather than module
identity. Predictions and ACT halting \citep{graves2016} are read from $\zH$.

Mechanistic interpretability studies neural computation through causal
mediators, circuits, and sparse features \citep{elhage2021,meng2022,bricken2023}.
Structured domains are useful because external geometry gives an interpretable
basis for evaluating internal variables: prior work finds spatially grounded
state in chess, maze-solving Transformers, and geographical representations
\citep{toshniwal2022,jenner2024,ivanitskiy2023,spies2025,desabbata2025}. Recent
spatial reasoning models extend this motivation to static VLMs, navigation, and
robot policies: SpatialVLM injects spatial supervision, NaviLLM-like models
tokenize viewpoint/history, and OpenVLA-style policies map vision-language state
to actions \citep{spatialvlm2024,navillm2024,openvla2024}. Interpretability for
these systems is moving from visualization toward probes, feature interventions,
spatial/temporal IDs, and reasoning traces \citep{spies2025,linearspatial2026,vilasr2025}.
These studies motivate a reusable framework: task-specific visualizations are
valuable, but a common notion of locality is needed to compare reasoning across
tasks and architectures.

For spatial reasoning, causal interpretability should ultimately ask what changes
when internal state is changed. Activation patching and related interventions
localize causal mediators in language models~\citep{meng2022,syed2023attribution},
while spatial-reasoning interpretability has begun to combine probes, sparse features,
and causal interventions in maze, grid-world, and VLM settings \citep{spies2025,linearspatial2026,vilasr2025}.
We therefore treat locality as an intervention-level question: after perturbing one
cell, object, or cycle state, does the effect remain near the source, stay inside
a semantic segment, or spread globally? Jacobian and attention kernels remain
useful structural summaries, but the main framework emphasizes finite
perturbations because they are closer to causal behavior under non-infinitesimal
changes.

\section{Experimental setup and methods}
\label{sec:setup}
\label{sec:methods}

We analyze released HRM and TRM checkpoints on Maze-Hard, Sudoku Extreme, and
ARC-AGI. HRM uses a $2\times2$ H/L schedule; TRM uses a shared module with
longer schedules (three high phases and four low calls for Maze/ARC-AGI, six for
Sudoku). Main analyses are anchored at critical cycle locations selected by
state-change convergence: HRM H1L1 for all tasks, TRM H2L3 for Maze/ARC-AGI,
and TRM H2L5 for Sudoku. Finite-noise patching uses $n=50$ examples per
model--task cell. ARC-AGI neighborhoods are same-color 4-connected foreground
components after padding grids to $30\times30$ and filtering to examples with
2--15 components. Appendix~\ref{app:jacobian} reports the structural Jacobian
analyses and their sample counts.

\paragraph{SAE semantic locality.}
For H and L activations at the critical cycle, we train independent sparse
autoencoders ($512\to2048$, $\lambda_1=10^{-3}$), rank features by ablation
impact, and compute how much each feature's effect remains within a task segment
(corridor, Sudoku house, or ARC object). SAE locality is human-readable but
feature-wise: it reveals where individual sparse features act, not necessarily
the aggregate topology of the whole update.

\paragraph{Finite-noise activation patching.}
For a source site $v$, injection level $a$, and capture level $b$, we perturb the
single activation vector $z_a[v]$ by $\sigma_a\epsilon$, with
$\epsilon\sim\mathcal{N}(0,I)$, and measure the activation-difference field at
each target site $u$:
\begin{equation}
  A_{a\to b}[u,v] = \|z_b^{\mathrm{patched}}[u]-z_b^{\mathrm{clean}}[u]\|_2,
  \qquad
  L_{a\to b}(v)=\frac{\sum_{u\in\mathcal{N}(v)}A_{a\to b}[u,v]}{\sum_u A_{a\to b}[u,v]}.
  \label{eq:finite_patch}
\end{equation}
The neighborhood $\mathcal{N}(v)$ is fixed before measurement: distance at most
one along the Maze path, the same $3\times3$ Sudoku subgrid, or the same
4-connected ARC-AGI foreground object. The denominator ranges over the same
valid task positions, and rows with zero total activation change are excluded
from the average. The random baseline is computed with the same neighborhood
sizes, so $L_{a\to b}$ is interpreted as locality above a geometry-matched null.
Noise scale is calibrated independently of the locality score: Maze and Sudoku
reuse the reliability-calibrated scale targeting roughly 30\% self-drop at a
probe position, while ARC-AGI uses the SNR=1 calibration from the cross-dataset
runs. We report within-L, within-H, and cross-level/cross-cycle analogs; higher
values mean the finite perturbation has more spatially local causal reach.
Structural Jacobian and attention analogs are defined and reported in
Appendix~\ref{app:jacobian} as topology checks rather than finite-causal effects.

\begin{table}[t]
  \centering
  \caption{Probe suite used to instantiate interaction locality. The main text
  prioritizes finite perturbations as causal evidence; SAE features provide
  semantic readability, and appendix-only structural probes report linearized
  topology and attention bias.}
  \label{tab:probe_suite}
  \footnotesize
  \setlength{\tabcolsep}{4pt}
  \begin{tabular}{@{}p{0.18\textwidth}p{0.29\textwidth}p{0.42\textwidth}@{}}
    \toprule
    Probe & Measurement & What it supports / limitation \\
    \midrule
    SAE features & Ablation impact within task segments & Human-readable feature locality; may miss distributed collective effects. \\
    Finite patching & Distance decay under source perturbations & Causal reach at finite noise scale, with reliability checks. \\
    Structural checks (appendix) & Jacobian and attention locality & Linearized or architectural topology, not the primary causal claim. \\
    \bottomrule
  \end{tabular}
\end{table}

\begin{figure*}[t]
  \centering
  \textbf{(a) Task geometries.}\\[-2pt]
  \includegraphics[width=0.94\textwidth]{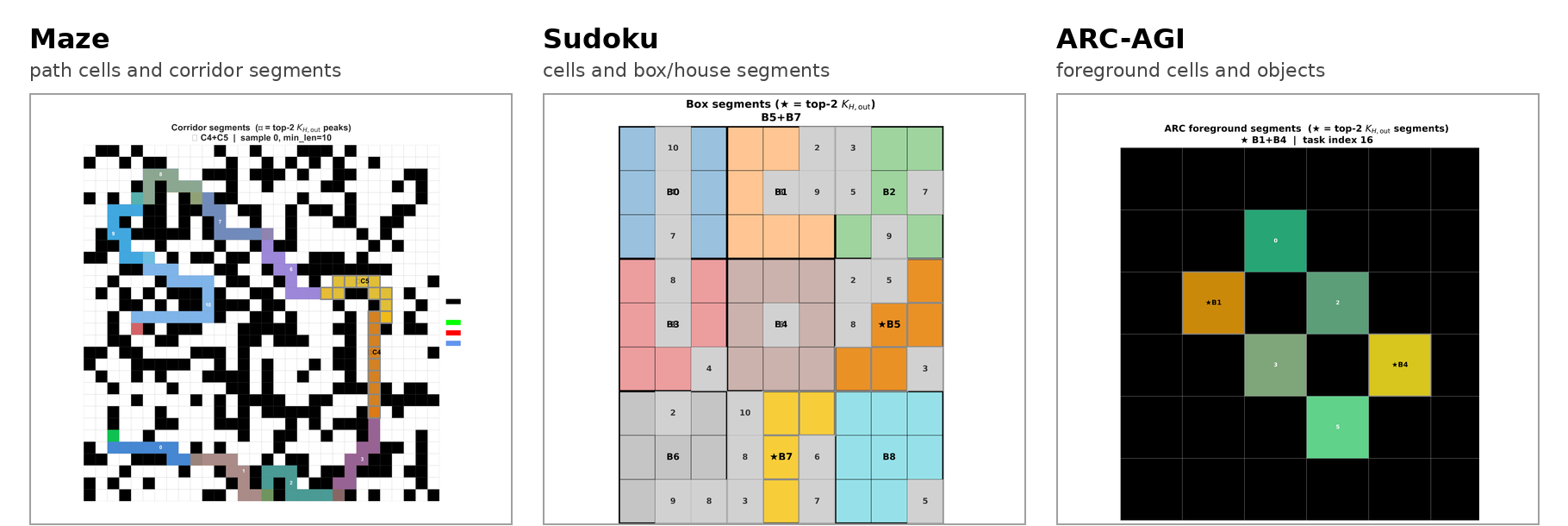}

  \vspace{0.35em}
  \textbf{(b) Interaction-locality pipeline.}\\[-2pt]
  \includegraphics[width=0.84\textwidth]{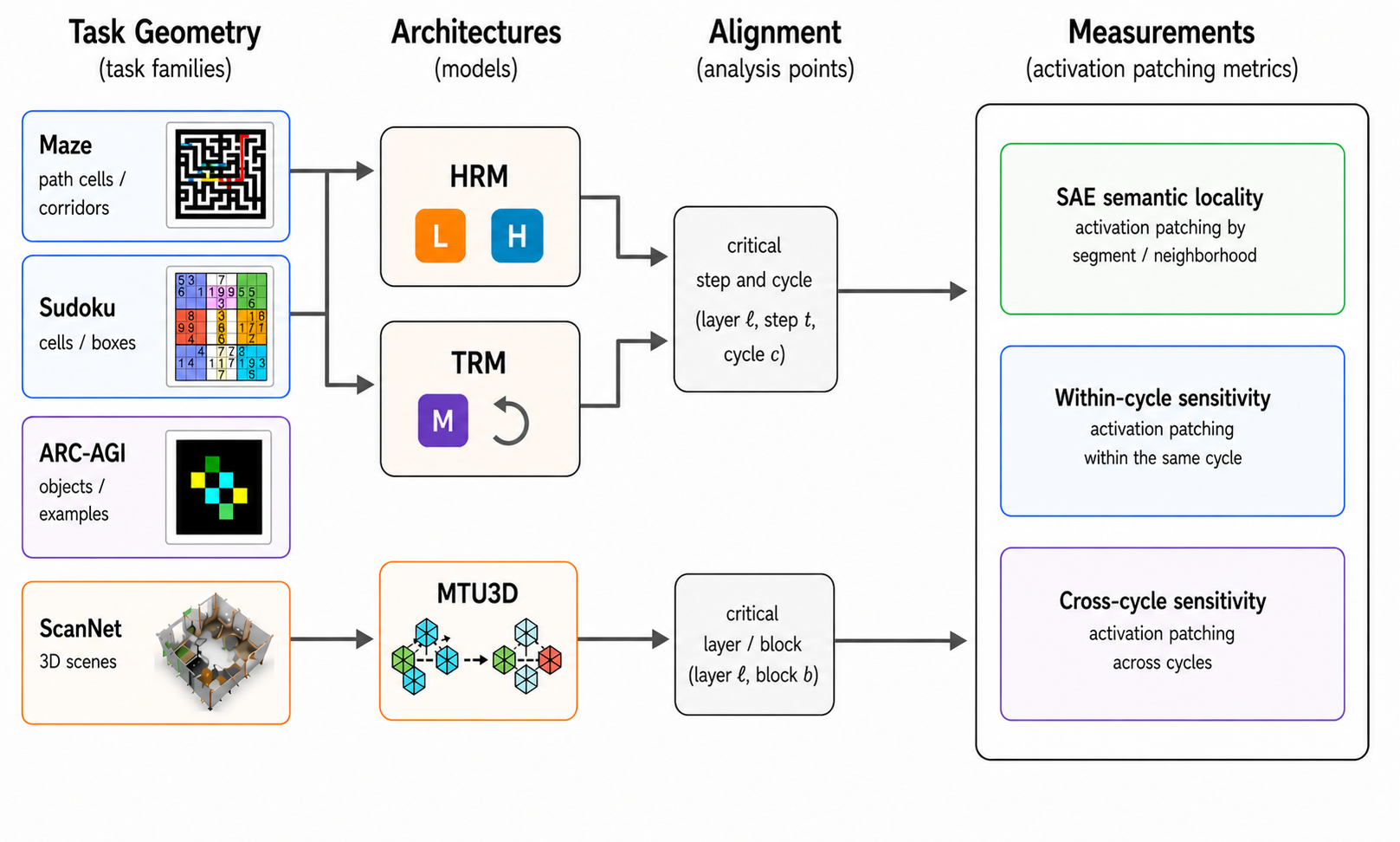}
  \caption{Interaction-locality framework. Panel (a) shows the task geometries
  that define locality: Maze path cells and corridors, Sudoku cells and
  $3\times3$ boxes, and ARC-AGI foreground objects/examples. Panel (b) shows
  how the same measurement suite is aligned across models: HRM and TRM are
  analyzed at a critical recursive step and cycle, while MTU3D/ScanNet is
  analyzed at critical layers or blocks. The green measurement box denotes SAE
  semantic locality by segment/neighborhood, the blue box denotes within-cycle
  or within-layer finite-noise patching sensitivity, and the purple box denotes
  cross-cycle or cross-layer patching sensitivity. Structural Jacobian and
  attention diagnostics are reported in the appendix as topology checks rather
  than the primary finite-intervention evidence.}
  \label{fig:overview}
\end{figure*}

\paragraph{MTU3D/ScanNet.}
For the embodied 3D extension, we analyze MTU3D \citep{mtu3d2025} on ScanNet indoor scenes
\citep{dai2017scannet}. MTU3D is a 3D occupancy-transformer model that predicts
scene structure from object-level scene features; ScanNet provides reconstructed
indoor environments with object geometry and semantic annotations. We evaluate
interaction locality on 30 ScanNet scenes by patching object-level activations
at selected layers/blocks and measuring changes in the model's scene-completion
or occupancy prediction metric. Local neighborhoods are defined by object
distance in the reconstructed 3D scene, and random near-object patches provide
the geometry-matched baseline.

\paragraph{Code availability.}
Our code is available online (\url{https://anonymous.4open.science/r/TinyMechInterp-6D42/}).



\section{Results}
\label{sec:results}

\subsection{Convergence identifies the analysis window}
\label{sec:convergence}

\begin{figure*}[t]
  \centering
  \includegraphics[width=0.90\textwidth]{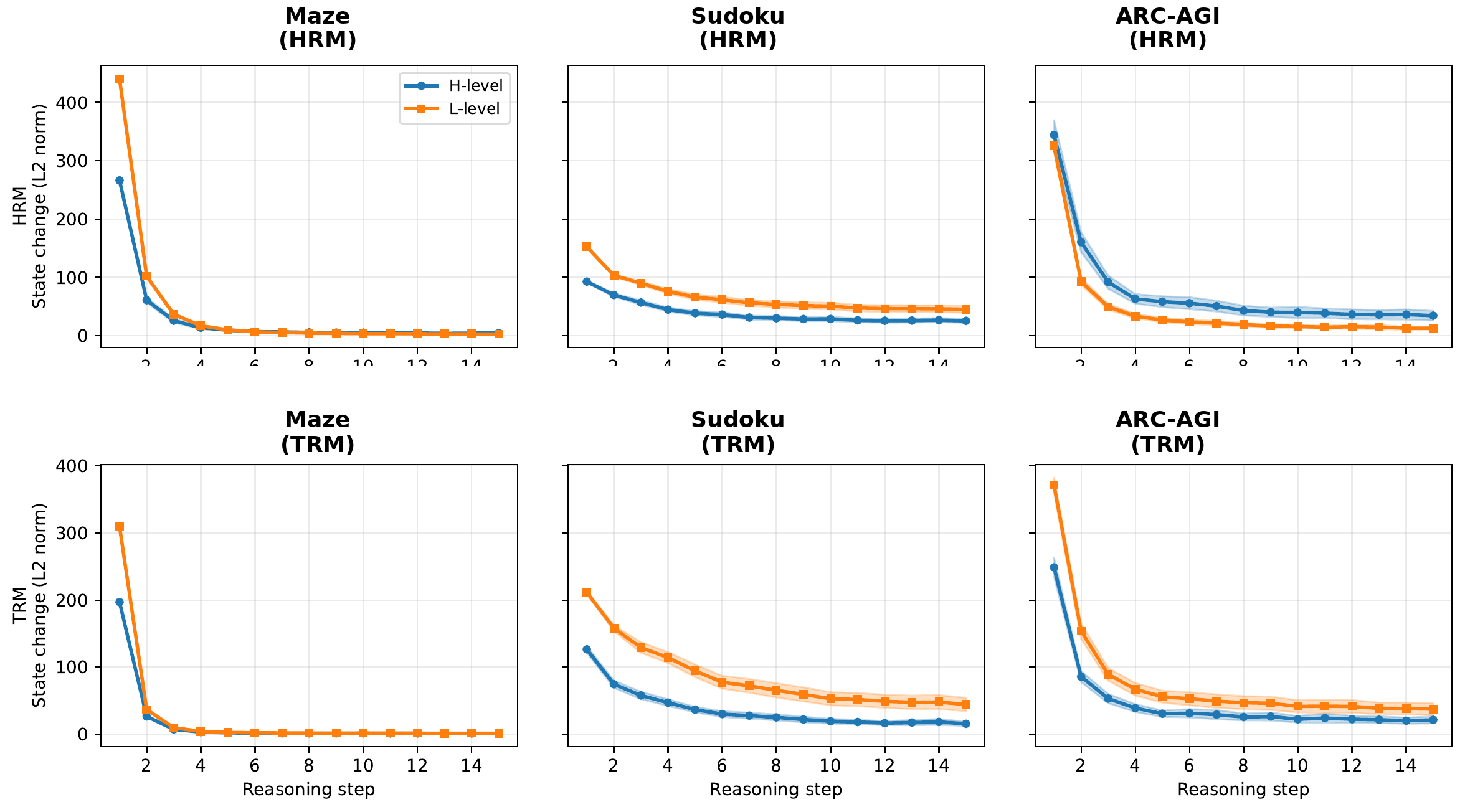}
  \caption{H/L hidden-state change across reasoning steps for HRM (top) and TRM
  (bottom). Blue curves track H-level changes and orange curves track L-level
  changes; shaded bands denote 95\% bootstrap CIs over examples. Maze changes
  sharply after the first step, Sudoku refines over several steps, and ARC-AGI
  lacks a single convergence cliff; these trajectories select the critical
  windows used by later analyses. \Cref{fig:qvalue} gives the Q-value view.}
  \label{fig:convergence}
\end{figure*}

\begin{figure*}[t]
  \centering
  \setlength{\tabcolsep}{1pt}
  \begin{tabular}{ccc}
    \includegraphics[width=0.30\textwidth]{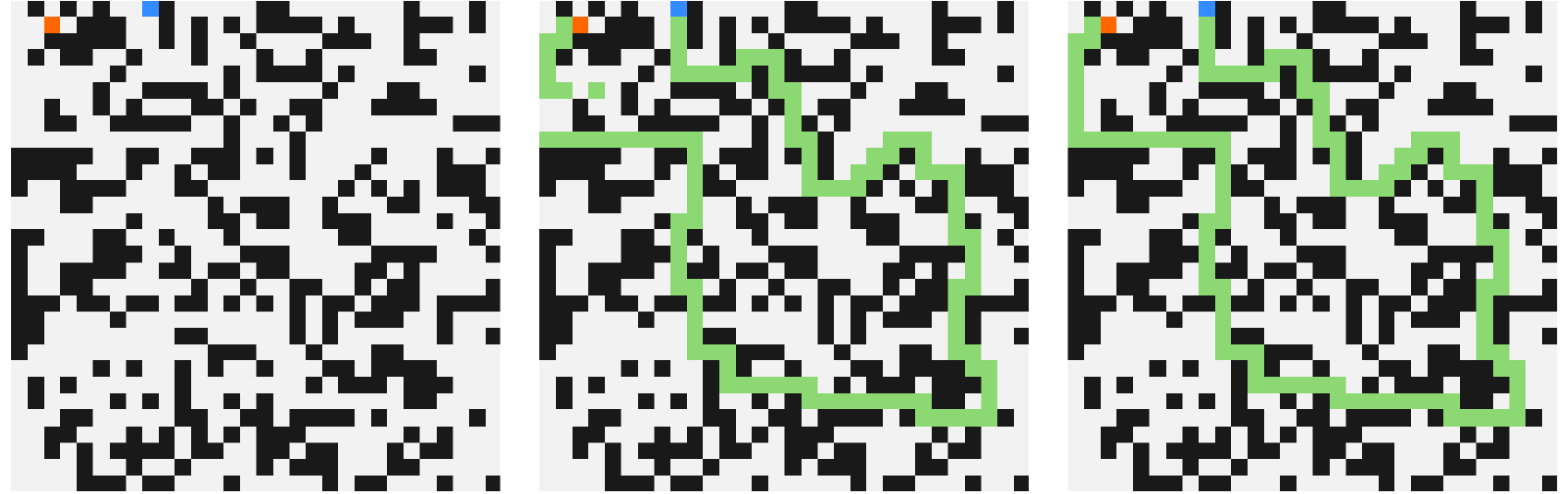} &
    \includegraphics[width=0.30\textwidth]{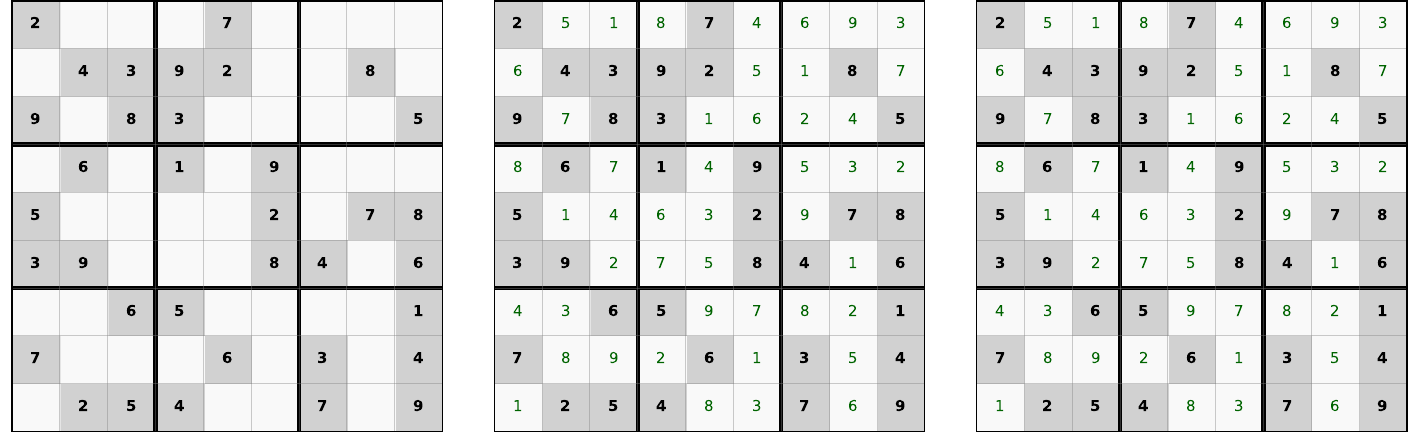} &
    \includegraphics[width=0.30\textwidth]{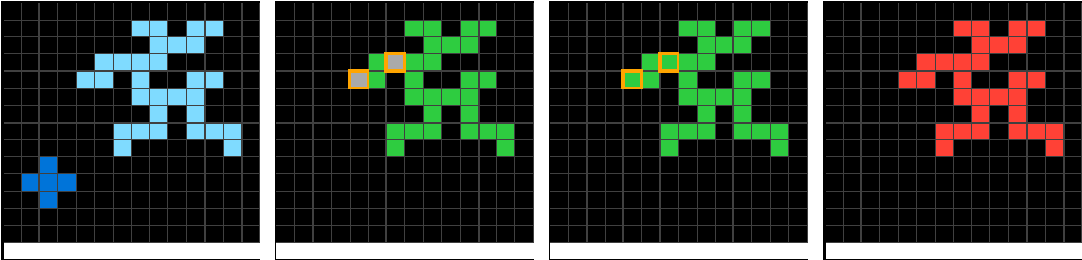} \\
    \includegraphics[width=0.30\textwidth]{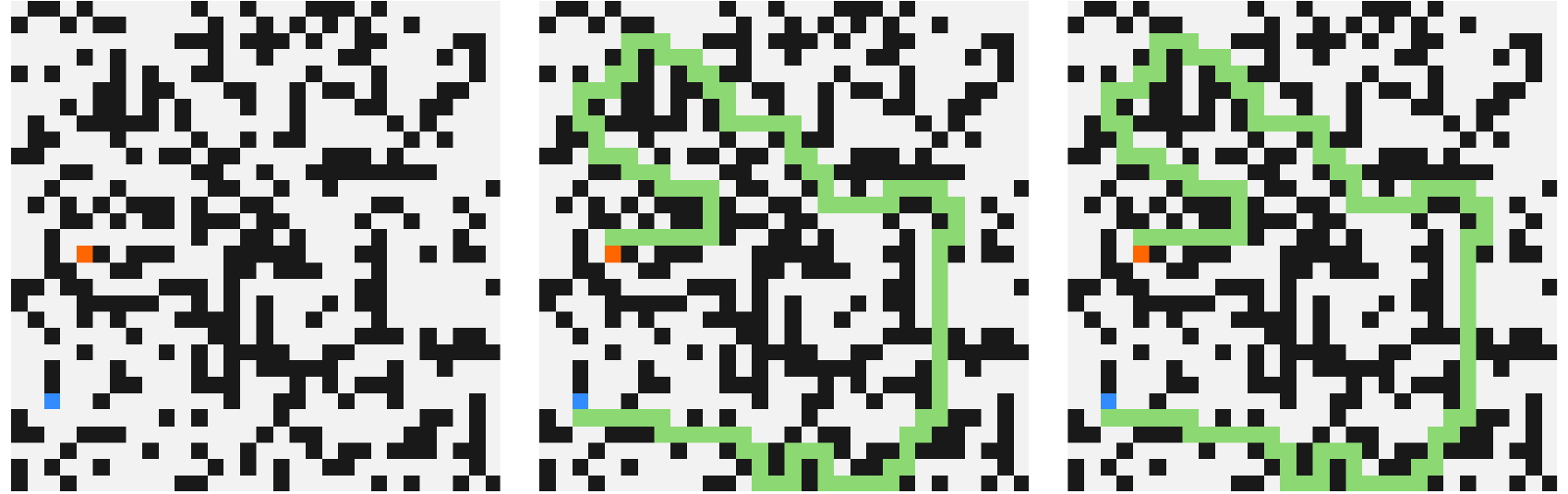} &
    \includegraphics[width=0.30\textwidth]{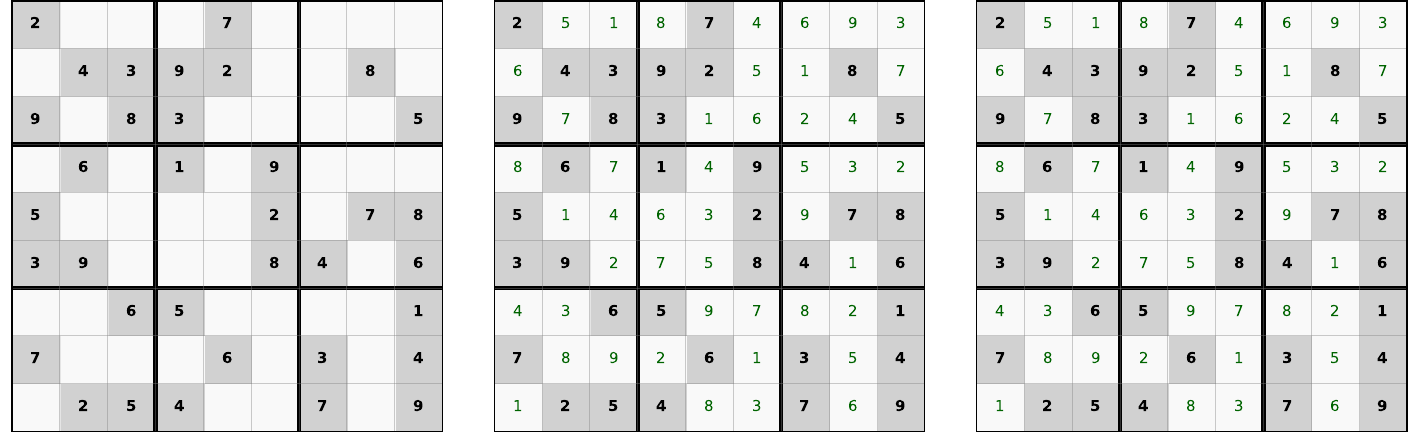} &
    \includegraphics[width=0.30\textwidth]{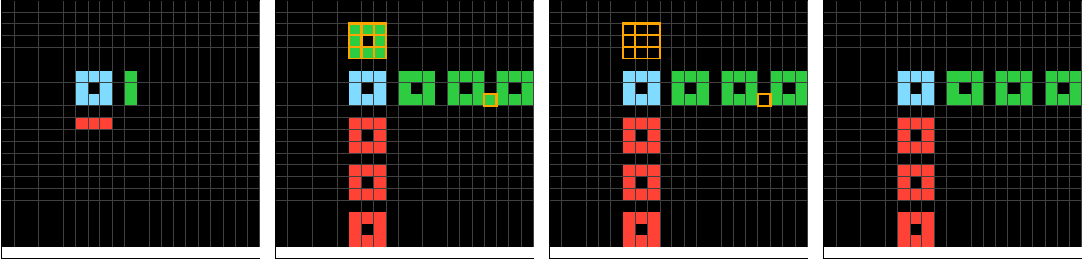}
  \end{tabular}
  \caption{Initial, step-1, and final decoded outputs for all model--task pairs
  (HRM top row, TRM bottom row; columns are Maze, Sudoku, ARC-AGI). Each panel
  shows the decoded task structure at the input/initial state, after the first
  reasoning step, and at the final step; colors follow the task-specific output
  convention for paths, digits, or ARC objects. Step 1 already contains
  recognizable task structure across domains, supporting it as a solution-forming
  window.}
  \label{fig:step1_final}
\end{figure*}

Hidden states change most sharply early in reasoning. \Cref{fig:convergence}
shows that Maze has the clearest first-step drop, Sudoku decays over multiple
steps, and ARC-AGI has no single cliff. The qualitative comparisons in
\Cref{fig:step1_final} show why this early window is worth analyzing: first-step
decodes already contain much of the final solution structure. At the same time,
these decodes are observational. They justify the analysis window but do not
establish which positions causally influence later updates. The remainder of the
paper therefore uses interaction kernels and finite perturbations for
mechanistic claims.

\subsection{Activation patching gives the locality fingerprint}
\label{sec:results_patching}
\label{sec:results_sae}

\begin{figure*}[t]
  \centering
  \begin{minipage}[b]{0.49\textwidth}
    \centering
    \includegraphics[width=\textwidth]{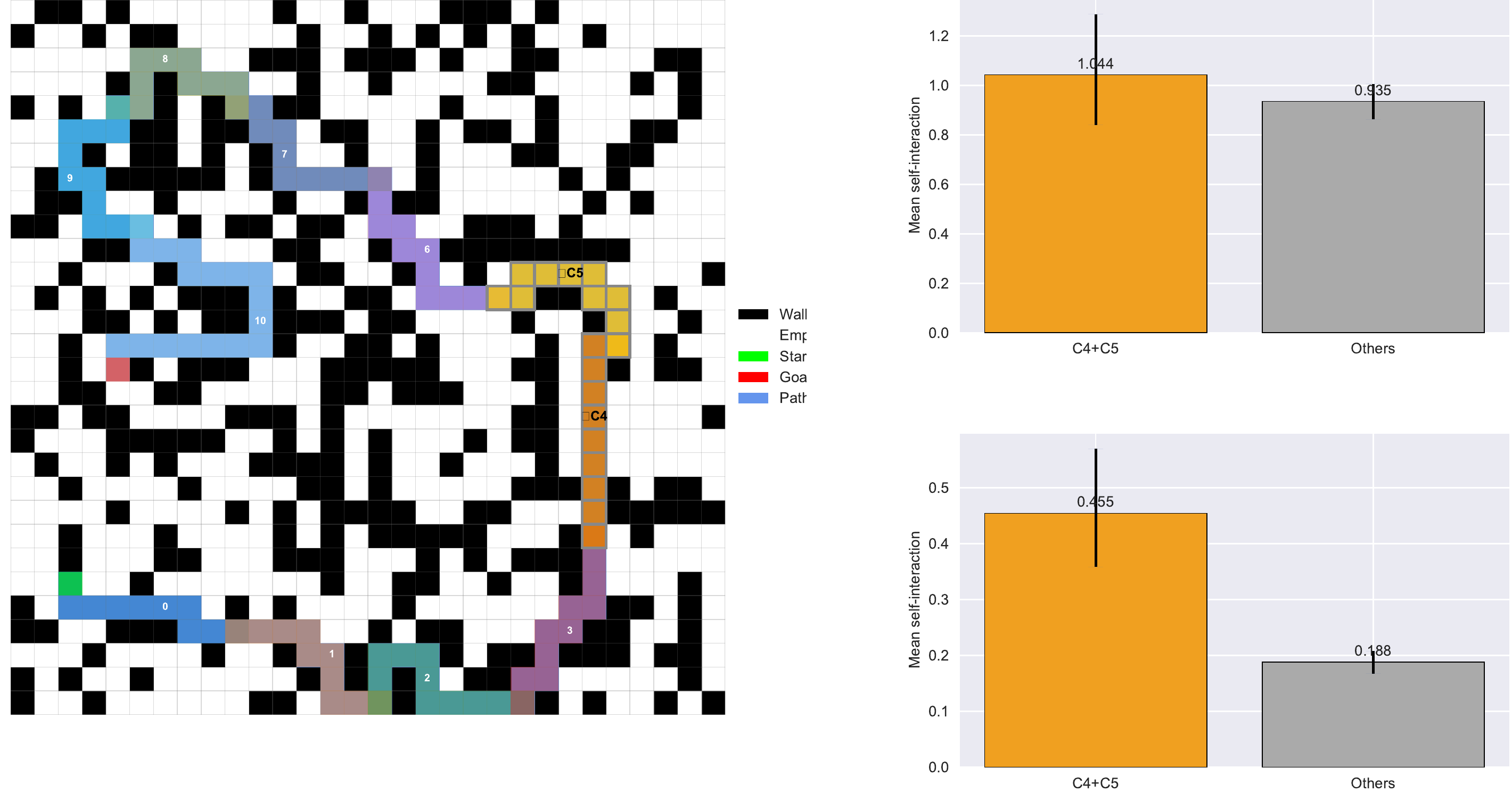}
  \end{minipage}\hfill
  \begin{minipage}[b]{0.49\textwidth}
    \centering
    \includegraphics[width=\textwidth]{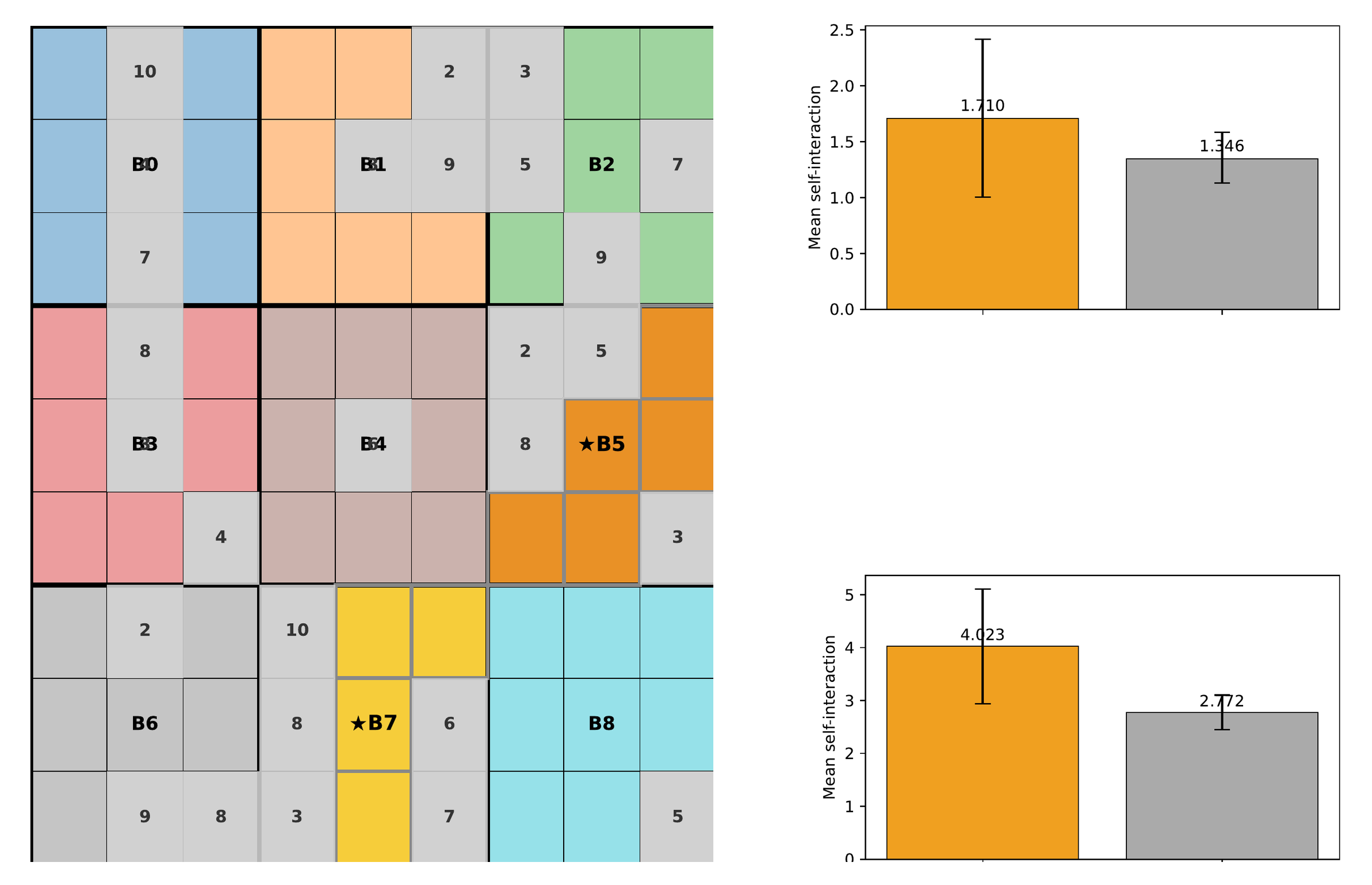}
  \end{minipage}
  \caption{Representative SAE semantic locality examples. Left: a Maze HRM
  feature whose ablation concentrates on a corridor pair; the colored corridor
  overlay marks the selected segment and the bar plots compare that segment with
  all other path cells. Right: a Sudoku HRM feature whose ablation concentrates
  on a pair of $3\times3$ boxes; orange bars denote the highlighted box pair and
  gray bars denote the comparison set, with error bars showing 95\% CIs. These
  feature ablations give interpretable segment-level hypotheses, such as
  corridors and Sudoku boxes; causal reach is tested by the finite-noise patching
  results in \Cref{tab:patching_summary,fig:activation_patching_main}.}
  \label{fig:top2_segments_main}
\end{figure*}

SAE features provide a useful segment-level view (Figure~\ref{fig:top2_segments_main}): Maze HRM has more local
L-level features than H-level features, while Sudoku and ARC-AGI features are
mostly balanced under the chosen segment definitions. This tells us where
individual sparse features act, but the causally stronger test is finite-noise
activation patching: after corrupting one source site, we measure where the
resulting hidden-state change travels.

\Cref{tab:patching_summary} gives the cross-task activation-patching summary.
The most stable pattern is within-H locality. Across all six model--task pairs,
within-H is at least as local as within-L and is above the corresponding
random-neighborhood baseline. The effect is clearest in TRM/Maze, where within-H
is .580 while within-L is near baseline at .032, and it extends to ARC-AGI when
connected objects define the neighborhood. Sudoku shows the same direction under
a larger $3\times3$ house neighborhood, where the locality unit already spans a
substantial constraint region.

\begin{table*}[t]
  \centering
  \caption{Finite-noise activation-patching locality summary (mean [95\% CI]).
  Locality is $L_{a\to b}$ from \Cref{eq:finite_patch}: the fraction of
  activation-change mass inside the task-defined neighborhood; baseline is the
  expected local fraction under the same neighborhood sizes. Within-H is at least
  as local as within-L in every model--task cell and remains above the
  geometry-matched baseline, supporting a consistent local-write signature under
  finite perturbations while allowing cross-cycle channels to vary with task and
  architecture.}
  \label{tab:patching_summary}
  \scriptsize
  \setlength{\tabcolsep}{3.2pt}
  \begin{tabular}{llccccc}
    \toprule
    Architecture & Task & $n$ & Baseline & within-L & within-H & cross-HH \\
    \midrule
    HRM & Maze    & 50 & .026 & .160 [.157,.162] & .373 [.361,.384] & .225 [.217,.233] \\
    HRM & Sudoku  & 30 & .111 & .371 [.368,.374] & .374 [.370,.377] & .338 [.335,.341] \\
    HRM & ARC-AGI & 30 & .384 & .556 [.501,.616] & .619 [.564,.677] & .550 [.491,.612] \\
    \midrule
    TRM & Maze    & 50 & .026 & .032 [.032,.032] & .580 [.567,.594] & .039 [.038,.040] \\
    TRM & Sudoku  & 30 & .111 & .154 [.153,.156] & .193 [.190,.196] & .210 [.203,.218] \\
    TRM & ARC-AGI & 30 & .314 & .625 [.577,.671] & .754 [.697,.805] & .434 [.363,.506] \\
    \bottomrule
  \end{tabular}
\end{table*}

\begin{figure*}[t]
  \centering
  \includegraphics[width=0.80\textwidth]{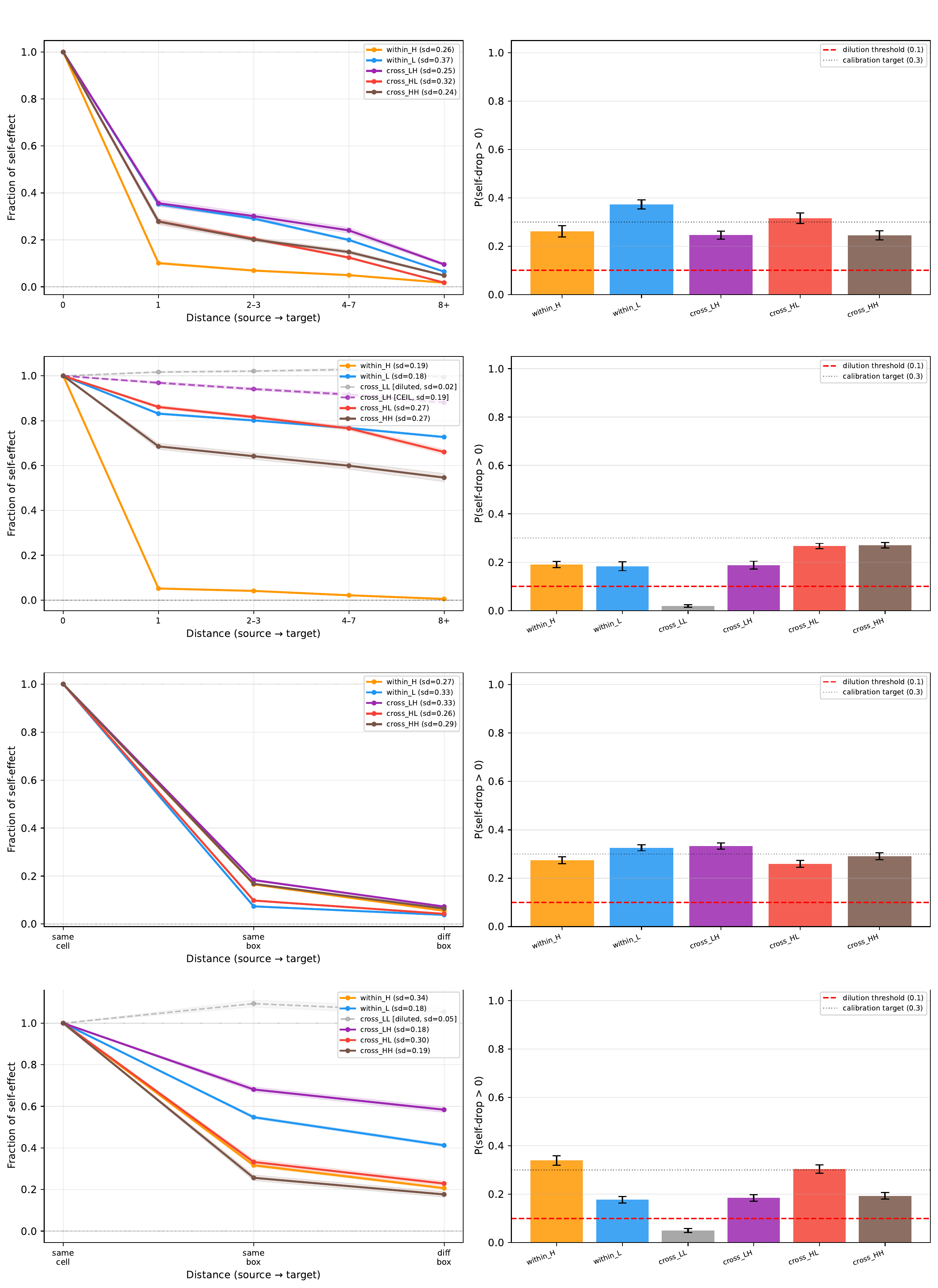}
  \caption{Finite-noise activation patching analogs for Maze and Sudoku. Left
  column: normalized spatial decay of the self-effect as the patch source moves
  from the target cell to farther cells, boxes, or segments. Right column:
  self-drop reliability. The gray dotted line at 0.3 is the self-drop calibration
  target used to set the perturbation scale, and the red dashed line at 0.1 marks
  a dilution threshold below which a channel has little effect on its own source
  prediction. Colors denote within-H, within-L, and cross-level/cross-cycle
  source--target channels. HRM/Maze separates local H writes from broader
  cross-cycle H-state propagation, while TRM/Maze has a flatter shared-operator
  channel. Full raw drops and heatmaps are in Appendix~\ref{app:patching}.}
  \label{fig:activation_patching_main}
\end{figure*}

\Cref{fig:activation_patching_main} shows why reliability diagnostics are needed.
In HRM/Maze, direct within-H perturbations are concentrated and reliable:
H-level writes local summaries at each token. Perturbations carried by $\zH$
across cycles decay more slowly, showing that the H state can propagate those
summaries more broadly. TRM/Maze contrasts with an almost baseline within-L
channel under the shared recursive operator. Sudoku has higher locality at both
levels because a $3\times3$ house already forms a broad constraint unit, but it
preserves the same within-H direction.

\subsection{MTU3D extends the framework to embodied 3D grounding}
\label{sec:results_mtu3d}

\begin{figure*}[t]
  \centering
  \includegraphicsmaybe[width=0.86\textwidth]{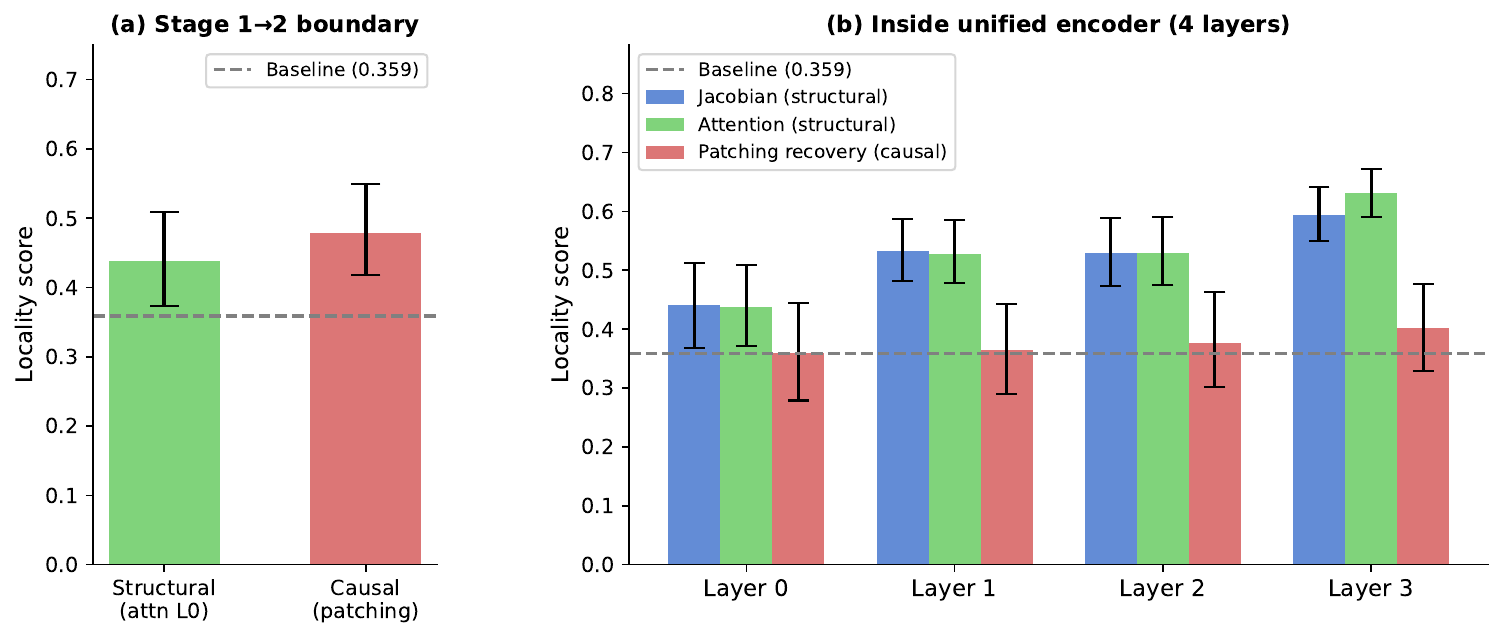}
  \caption{Interaction locality on MTU3D/ScanNet. Bars show locality scores with
  95\% CIs; the dashed horizontal line is the random near-object baseline computed
  from the same ScanNet object-distance neighborhoods. Panel (a) evaluates the
  visual-to-grounding handoff by comparing layer-0 structural attention locality
  with causal input patching after corrupting one object feature. Panel (b)
  compares structural Jacobian locality, attention locality, and layer-recovery
  patching across the four unified-encoder layers. In this MTU3D/ScanNet setting,
  structural (layer-0 attention, .438) and causal (input patching, .479) locality
  both exceed the random baseline (.359) at the handoff, but layer-recovery
  patching inside the encoder is near baseline despite structural
  Jacobian/attention locality above it. The dissociation separates architectural
  spatial bias from causal local recovery in this 3D embodied setting.}
  \label{fig:mtu3d_main}
\end{figure*}

We stress-test the framework on MTU3D, a large-scale 3D embodied scene-grounding
model \citep{mtu3d2025}, using ScanNet indoor scenes \citep{dai2017scannet}.
The experiment asks whether corrupting or restoring one object-level 3D scene
feature changes nearby objects more than distant objects. Thus MTU3D operates on
metric 3D scene objects rather than 2D grid cells, and neighborhoods are defined
by object distance in the ScanNet scene. \Cref{fig:mtu3d_main} reveals two distinct outcomes. Panel (a) shows
converging evidence at the visual-to-grounding handoff (the MTU3D Stage 1--2
boundary): structural layer-0 attention locality (.438) and causal input patching
(.479) both exceed the random baseline
(.359), confirming the framework's sensitivity when spatial structure is present.
Yet layer-recovery patching inside the unified encoder remains at baseline
throughout all four layers, even though structural Jacobian and attention scores
are well above baseline (panel b). The dissociation is itself informative in this MTU3D setting: MTU3D has a clear
spatial inductive bias in its architecture, but this bias does not compound into
causal locality through encoder depth. Together with the HRM/TRM results, this
suggests that the observed local-to-global handoff is tied to recursive reasoning
dynamics rather than being a generic consequence of spatial attention alone.

\subsection{Triangulating claims across probes}
\label{sec:triangulation}

Appendix~\Cref{tab:triangulation} summarizes how the probes jointly support the paper's
main claims. The organization is layered: finite-noise patching supplies the
main causal evidence, SAE feature ablations make the task segments readable, and
appendix structural checks explain the corresponding linearized topology. This
triangulation makes the interaction-locality framework interpretable at both the
semantic-feature level and the finite-intervention level.

\section{Discussion}
\label{sec:discussion}

Interaction locality, the main object of study, unifies semantic feature analysis and finite causal perturbations under the same
task geometry. This improves clarity and falsifiability: a claim such as \textit{H is
global} becomes a statement about which intervention channel is local, which
channel propagates broadly, and which neighborhood defines locality. The results
therefore revise the informal H/L story. H-level states can write locally within
a cycle, especially under activation patching, while still carrying those local
summaries across recursive time.

A key implication is that local and global computation should not be treated as
static properties of a named state. They are relations among a state, a task
geometry, a perturbation scale, and a recursive time edge. SAE ablations give
readable hypotheses about which corridors, boxes, or objects a sparse feature
affects, but do not by themselves establish causal reach. Finite-noise patching
checks whether the same locality survives a non-infinitesimal intervention, but
requires reliability diagnostics to avoid mistaking signal dilution for broad
influence. The appendix Jacobian and attention analyses add structural context,
but the main claims rely on intervention-level evidence.

The MTU3D extension clarifies the framework's scope. In the 30-scene
MTU3D/ScanNet setting, structural spatial bias is present in attention and
Jacobian scores, yet causal recovery inside the encoder is near baseline. This
dissociation is precisely why interaction locality should compare multiple
probes: an architecture can attend locally without making downstream causal
recovery local. Conversely, HRM/TRM suggest that recursive cycles can create a
local-to-global handoff even in compact toy benchmarks. The framework is
therefore not tied to Maze, Sudoku, or ARC-AGI; it asks which task geometry makes
a causal interaction local.

For model design, the results suggest locality-aware objectives. A model could
receive separate losses for local move direction, segment-to-segment matching,
and cycle-level consistency, or regularizers that encourage the right
intervention scale at the right recursive phase. In navigation this might mean
supervising local collision-avoidance separately from route-segment matching; in
manipulation it might mean separating contact-local corrections from object-level
goal relations. Such objectives are not evaluated here, but interaction locality
gives an operational diagnostic for whether they work: after training, the
intended local and global channels should appear in the corresponding feature,
patching, structural, and temporal measurements.

\section{Limitations and future work}
\label{sec:limitations}

The analysis compares released checkpoints rather than an accuracy-matched
training sweep, and ARC-AGI uses HRM on ARC-AGI-2 and TRM on ARC-AGI-1. We
therefore make claims about mechanisms observed in available compact reasoning
models, not about which architecture is more accurate. The framework is also
geometry-dependent: our neighborhoods are appropriate for Maze, Sudoku, ARC-AGI,
and the MTU3D object-distance analysis, but they do not exhaust all possible
spatial relations. Sudoku also contains row and column constraints, ARC-AGI tasks
may require transformations across same-color object boundaries, and 3D embodied
models require choices about metric distance, contact, viewpoint, and
uncertainty. A first step toward the Sudoku row/column dimension is in \Cref{app:constraint_breakdown}.

A first future direction is training dynamics. If interaction locality is a
mechanistic signature rather than a post-hoc statistic, it should emerge, split,
or reorganize over checkpoints. Tracking SAE features, finite-noise reach,
structural kernels, and temporal propagation during training would reveal whether
local moves appear before global segment planning, whether H-boundary broadening
coincides with accuracy jumps, and whether TRM's shared-operator locality emerges
gradually or only after stable iterative routines appear. Such experiments would
also test whether locality-aware training objectives improve sample efficiency or
robustness.

A second direction is embodied spatial reasoning. MTU3D is only a first step:
real agents operate in 3D or 4D geometries with depth, contact, viewpoint, time,
and uncertainty. Extending interaction locality to these domains requires
replacing grid neighborhoods with scene graphs, contact graphs,
egocentric/allocentric coordinate frames, and time-varying object tracks. The
payoff could be substantial: for safety-critical navigation or manipulation, one
wants to know whether a local action is causally grounded in the intended global
plan, not merely whether the decoded rationale is plausible.

\section{Conclusion}
\label{sec:conclusion}

We introduced interaction locality as a task-geometry-aware interpretability
framework for recursive spatial reasoning. Across Maze-Hard, Sudoku Extreme, and
ARC-AGI, finite-noise activation patching gives a causally stronger account than
visualization alone: H-level writes tend to occupy the local same-segment channel
under task-defined neighborhoods, with the small HRM/Sudoku gap illustrating that
the magnitude of the H/L contrast depends on the chosen task scale. Cross-cycle
channels can still propagate those summaries broadly. SAE features provide readable segment-level
hypotheses, and appendix structural kernels explain the corresponding linearized
topology, but the main framework is grounded in finite interventions.

The broader contribution is a shift in what is measured. Instead of asking only
whether a compact recursive model has an interpretable feature or a plausible
decoded plan, the framework asks whether the information needed for that plan
moves at the right spatial scale and at the right recursive time. The MTU3D
extension shows why this distinction matters beyond puzzles: a large 3D embodied
model can exhibit structural spatial bias while lacking causal locality inside
its encoder. Interaction locality therefore becomes a diagnostic for whether
local actions, object-level relations, and global plans are causally connected,
not merely co-visible in a representation.

In this paper, Maze, Sudoku, ARC-AGI, and MTU3D provide geometries with known
neighborhoods. The same principle extends naturally to richer 3D and 4D settings
if the neighborhood relation is replaced by contact graphs, object tracks, scene
graphs, or egocentric/allocentric coordinate frames. The immediate future work is
therefore to track how interaction locality emerges during training and to test
whether locality-aware objectives improve robustness and transfer. If successful,
interaction locality could become a common diagnostic for spatial reasoning
models across symbolic puzzles, navigation, and embodied manipulation.

\clearpage
\bibliographystyle{plainnat}
\bibliography{tiny_mech_interp}

@article{hrm2024,
  author  = {Guan Wang and Jin Li and Yuhao Sun and Xing Chen and Changling Liu and Yue Wu and Meng Lu and Sen Song and Yasin Abbasi Yadkori},
  title   = {Hierarchical Reasoning Model},
  journal = {arXiv preprint arXiv:2506.21734},
  year    = {2025},
}

@article{trm2024,
  author  = {Alexia Jolicoeur-Martineau},
  title   = {Less is More: Recursive Reasoning with Tiny Networks},
  journal = {arXiv preprint arXiv:2510.04871},
  year    = {2025},
}

@article{graves2016,
  author  = {Alex Graves},
  title   = {Adaptive Computation Time for Recurrent Neural Networks},
  journal = {arXiv preprint arXiv:1603.08983},
  year    = {2016},
}

@article{elhage2021,
  author  = {Nelson Elhage and Neel Nanda and Catherine Olsson and Tom Henighan and Nicholas Joseph and Ben Mann and Amanda Askell and Yuntao Bai and Anna Chen and Tom Conerly and others},
  title   = {A Mathematical Framework for Transformer Circuits},
  journal = {Transformer Circuits Thread},
  year    = {2021},
}

@inproceedings{meng2022,
  author    = {Kevin Meng and David Bau and Alex Andonian and Yonatan Belinkov},
  title     = {{ROME}: Locating and Editing Factual Associations in {GPT}},
  booktitle = {Advances in Neural Information Processing Systems},
  year      = {2022},
}

@article{bricken2023,
  author  = {Trenton Bricken and others},
  title   = {Towards Monosemanticity: Decomposing Language Models with Dictionary Learning},
  journal = {Transformer Circuits Thread},
  year    = {2023},
}

@inproceedings{toshniwal2022,
  author    = {Shubham Toshniwal and others},
  title     = {Chess as a Testbed for Language Model State Tracking},
  booktitle = {Proceedings of the AAAI Conference on Artificial Intelligence},
  year      = {2022},
}

@inproceedings{jenner2024,
  author    = {Erik Jenner and others},
  title     = {Evidence of Learned Look-Ahead in a Chess-Playing Neural Network},
  booktitle = {Advances in Neural Information Processing Systems},
  year      = {2024},
}

@article{ivanitskiy2023,
  author  = {Michael I. Ivanitskiy and others},
  title   = {Structured World Representations in Maze-Solving Transformers},
  journal = {arXiv preprint arXiv:2312.02566},
  year    = {2023},
}

@article{desabbata2025,
  author  = {Stef De Sabbata and Stefano Mizzaro and Kevin Roitero},
  title   = {Geospatial Mechanistic Interpretability of Large Language Models},
  journal = {arXiv preprint arXiv:2505.03368},
  year    = {2025},
}

@article{spies2025,
  author  = {Alex F. Spies and William Edwards and Michael I. Ivanitskiy and Adrians Skapars and Tilman R\"auker and Katsumi Inoue and Alessandra Russo and Murray Shanahan},
  title   = {Transformers Use Causal World Models in Maze-Solving Tasks},
  journal = {arXiv preprint arXiv:2412.11867},
  year    = {2024},
}

@article{syed2023attribution,
  title         = {Attribution Patching Outperforms Automated Circuit Discovery},
  author        = {Syed, Aaquib and Rager, Can and Conmy, Arthur},
  journal       = {arXiv preprint arXiv:2310.10348},
  year          = {2023},
  eprint        = {2310.10348},
  archivePrefix = {arXiv},
  primaryClass  = {cs.LG},
  url           = {https://arxiv.org/abs/2310.10348}
}

@inproceedings{spatialvlm2024,
  title     = {{SpatialVLM}: Endowing Vision-Language Models with Spatial Reasoning Capabilities},
  author    = {Chen, Boyuan and Xu, Zhuo and Kirmani, Sean and Ichter, Brian and Driess, Danny and Florence, Pete and Sadigh, Dorsa and Guibas, Leonidas and Xia, Fei},
  booktitle = {Proceedings of the IEEE/CVF Conference on Computer Vision and Pattern Recognition},
  pages     = {14455--14465},
  year      = {2024},
  doi       = {10.1109/CVPR52733.2024.01370},
  url       = {https://arxiv.org/abs/2401.12168}
}

@inproceedings{navillm2024,
  title     = {Towards Learning a Generalist Model for Embodied Navigation},
  author    = {Zheng, Duo and Huang, Shijia and Zhao, Lin and Zhong, Yiwu and Wang, Liwei},
  booktitle = {Proceedings of the IEEE/CVF Conference on Computer Vision and Pattern Recognition},
  year      = {2024},
  doi       = {10.1109/CVPR52733.2024.01293},
  url       = {https://arxiv.org/abs/2312.02010}
}

@article{openvla2024,
  title         = {{OpenVLA}: An Open-Source Vision-Language-Action Model},
  author        = {Kim, Moo Jin and Pertsch, Karl and Karamcheti, Siddharth and Xiao, Ted and Balakrishna, Ashwin and Nair, Suraj and Rafailov, Rafael and Foster, Ethan and Lam, Grace and Sanketi, Pannag and Vuong, Quan and Kollar, Thomas and Burchfiel, Benjamin and Tedrake, Russ and Sadigh, Dorsa and Levine, Sergey and Liang, Percy and Finn, Chelsea},
  journal       = {arXiv preprint arXiv:2406.09246},
  year          = {2024},
  eprint        = {2406.09246},
  archivePrefix = {arXiv},
  url           = {https://arxiv.org/abs/2406.09246}
}

@article{vilasr2025,
  title         = {Reinforcing Spatial Reasoning in Vision-Language Models with Interwoven Thinking and Visual Drawing},
  author        = {Wu, Junfei and Guan, Jian and Feng, Kaituo and Liu, Qiang and Wu, Shu and Wang, Liang and Wu, Wei and Tan, Tieniu},
  journal       = {arXiv preprint arXiv:2506.09965},
  year          = {2025},
  eprint        = {2506.09965},
  archivePrefix = {arXiv},
  url           = {https://arxiv.org/abs/2506.09965}
}

@inproceedings{linearspatial2026,
  title     = {Linear Mechanisms for Spatiotemporal Reasoning in Vision Language Models},
  author    = {Kang, Raphaela and Chen, Hongqiao and Gkioxari, Georgia and Perona, Pietro},
  booktitle = {International Conference on Learning Representations},
  year      = {2026},
  url       = {https://arxiv.org/abs/2601.12626}
}

@incollection{vaswani2017attention,
title = {Attention is All you Need},
author = {Vaswani, Ashish and Shazeer, Noam and Parmar, Niki and Uszkoreit, Jakob and Jones, Llion and Gomez, Aidan N and Kaiser, \L ukasz and Polosukhin, Illia},
booktitle = {Advances in Neural Information Processing Systems 30},
pages = {5998--6008},
year = {2017},
publisher = {Curran Associates, Inc.}
}

@article{mtu3d2025,
  title   = {Move to Understand a 3D Scene: Bridging Visual Grounding and Exploration for Efficient and Versatile Embodied Navigation},
  author  = {Zhu, Ziyu and Wang, Xilin and Li, Yixuan and Zhang, Zhuofan and Ma, Xiaojian and Chen, Yixin and Jia, Baoxiong and Liang, Wei and Yu, Qian and Deng, Zhidong and Huang, Siyuan and Li, Qing},
  journal = {International Conference on Computer Vision (ICCV)},
  year    = {2025},
  url     = {https://mtu3d.github.io/}
}

@inproceedings{dai2017scannet,
    title={ScanNet: Richly-annotated 3D Reconstructions of Indoor Scenes},
    author={Dai, Angela and Chang, Angel X. and Savva, Manolis and Halber, Maciej and Funkhouser, Thomas and Nie{\ss}ner, Matthias},
    booktitle = {Proc. Computer Vision and Pattern Recognition (CVPR), IEEE},
    year = {2017}
}
\clearpage
\onecolumn
\raggedbottom

\appendix

\section{Claim-level triangulation details}
\label{app:triangulation}

\begin{table*}[t]
  \centering
  \caption{Claim-level triangulation. The framework
  connects semantic readability, finite-intervention locality, and structural
  topology so each mechanistic claim has a dedicated source of evidence.}
  \label{tab:triangulation}
  \footnotesize
  \setlength{\tabcolsep}{3pt}
  \resizebox{\textwidth}{!}{%
  \begin{tabular}{@{}p{0.24\textwidth}p{0.36\textwidth}p{0.32\textwidth}@{}}
    \toprule
    Mechanistic claim & Main evidence & Supporting detail \\
    \midrule
    Step 1 is a meaningful analysis window & Hidden-state drops and step-1 decodes show early solution structure & Patching and structural kernels analyze the causal interactions inside this window. \\
    SAE features provide readable segment hypotheses & Top features align with corridors and Sudoku boxes in representative cases & Full feature panels and task-level segment decompositions are in Appendix~\ref{app:sae}. \\
    H-level writes tend to use the local same-segment channel & Activation patching shows within-H $\ge$ within-L point estimates across all six HRM/TRM task cells & The small HRM/Sudoku gap is interpreted as a scale-selection effect because a Sudoku house is already a broad local constraint unit. \\
    Recursive H state can propagate summaries more broadly & Maze patching separates local within-H writes from broader cross-cycle H-state effects & Raw drops and heatmaps are reported in Appendix~\ref{app:patching}. \\
    MTU3D separates spatial bias from causal locality in this 30-scene setting & Stage 1--2 input patching is local, but encoder recovery patching is at baseline & Checkpoint and attention-distance controls are in Appendix~\ref{app:mtu3d}. \\
    Structural Jacobian topology contextualizes the main claim & Appendix kernels recover linearized cell/segment topology & These diagnostics show how the same task geometry appears in local linear interactions. \\
    \bottomrule
  \end{tabular}%
  }
\end{table*}

\section{Experimental Details}
\paragraph{MTU3D/ScanNet setting.}
For the 3D embodied extension, MTU3D is the model and ScanNet is the task
source: we analyze precomputed object-level Stage-1 features from 30 ScanNet
indoor scenes \citep{dai2017scannet}. The measured task is scene grounding over
3D object/voxel representations. Locality neighborhoods are defined by metric
object distance, calibrated so that the near set contains roughly seven nearest
neighbors per scene; the random baseline is the corresponding fraction of
near-object off-diagonal pairs. The MTU3D experiments therefore test whether
input corruption or layer recovery changes spatially nearby scene objects more
than distant objects.

\paragraph{Compute resources.}
All reported analyses use released checkpoints and do not require retraining the
HRM/TRM models. The accompanying code is written in PyTorch and is intended to
run on a single CUDA GPU; the repository recommends a GPU with at least 6GB VRAM
for notebook-level analyses, with selective state capture and batch size 1 used
for memory-intensive activation and intervention runs. The setup script
downloads dependencies, datasets, and checkpoints; checkpoint sizes range from
tens of MB for small TRM checkpoints to approximately 2GB for ARC-AGI
checkpoints. The finite-noise activation-patching, SAE-ablation, and Jacobian
diagnostic scripts process examples independently, so runtime scales
approximately linearly with the number of evaluated samples. Reproducing the
optional Maze-TRM training run from scratch is documented as requiring roughly
40 hours on a single A100 40GB GPU, but the paper's main experiments use the
released/analyzed checkpoints rather than retraining them. 

\section{Additional discussion}\label{app:disc}
\paragraph{Broader impacts.}
Interaction locality is intended as an auditing tool for spatial reasoning
models. Its positive impact is to make compact recursive and embodied spatial
systems more inspectable before they are used in safety-relevant settings such
as navigation, manipulation, and scene grounding. The same analysis could also
inform stronger spatial reasoning systems, including systems deployed in
physical environments. If such systems are used without adequate safety checks,
errors in locality measurement or overconfidence in interpretability results
could contribute to unsafe navigation, manipulation failures, or misplaced trust
in embodied agents. We therefore view interaction-locality analysis as a
diagnostic complement to, not a replacement for, task-specific safety,
robustness, and deployment evaluations.
\paragraph{Licence.}
The official HRM implementation is released under Apache-2.0 and links to the
official Sapient checkpoint artifacts used in this work. The checkpoint pages are
publicly released by the same authors, but do not expose separate license
metadata beyond the official repository release context.

\section{Additional convergence results}
\label{app:convergence}

\begin{center}
  \includegraphics[width=0.86\textwidth]{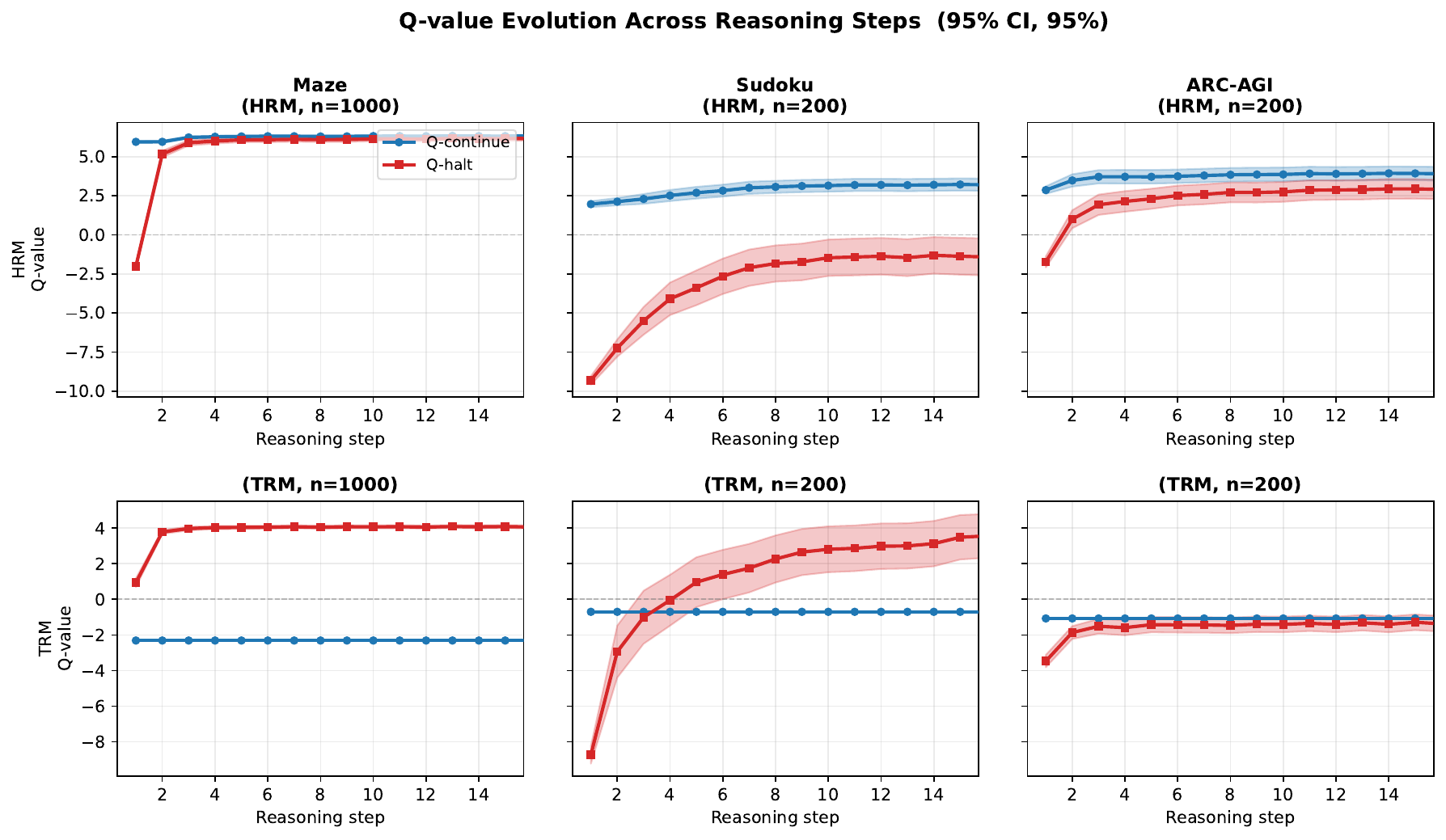}
  \captionof{figure}{Q-value evolution across reasoning steps for HRM and TRM on
    all three tasks. The scalar Q-value trajectories
    support the same critical-window selection as hidden-state divergence: Maze
    changes early, Sudoku refines over more steps, and ARC-AGI lacks a single
    sharp cliff.}
  \label{fig:qvalue}
\end{center}

\Cref{fig:qvalue} gives the scalar view corresponding to the hidden-state
changes in \Cref{fig:convergence}. Cycle-level diagnostics support the
critical-cycle choices reported in Section~\ref{sec:setup}.

\section{Additional first-step cycle decodes}
\label{app:cycle_decodes}

\begin{center}
  \begin{minipage}{0.40\textwidth}
    \centering
    \textbf{Maze HRM: remaining L-cycle decodes}\par\vspace{2pt}
    \includegraphics[width=\textwidth]{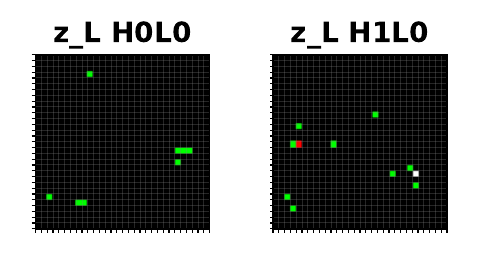}
  \end{minipage}\hfill
  \begin{minipage}{0.52\textwidth}
    \centering
    \textbf{Maze TRM: remaining upstream $z_L$ decodes}\par\vspace{2pt}
    \includegraphics[width=\textwidth]{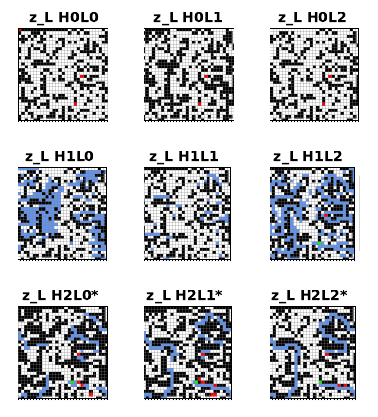}
  \end{minipage}
  \captionof{figure}{Maze cycle-decode panels not shown in the main selected figure.
    These additional observational decodes restore the first-step cycle context
    while keeping the main text focused on H-update endpoints.}
  \label{fig:app_cycle_decode_maze_remaining}
\end{center}

\begin{center}
  \textbf{HRM / Sudoku}\par\vspace{2pt}
  \includegraphics[width=0.72\textwidth]{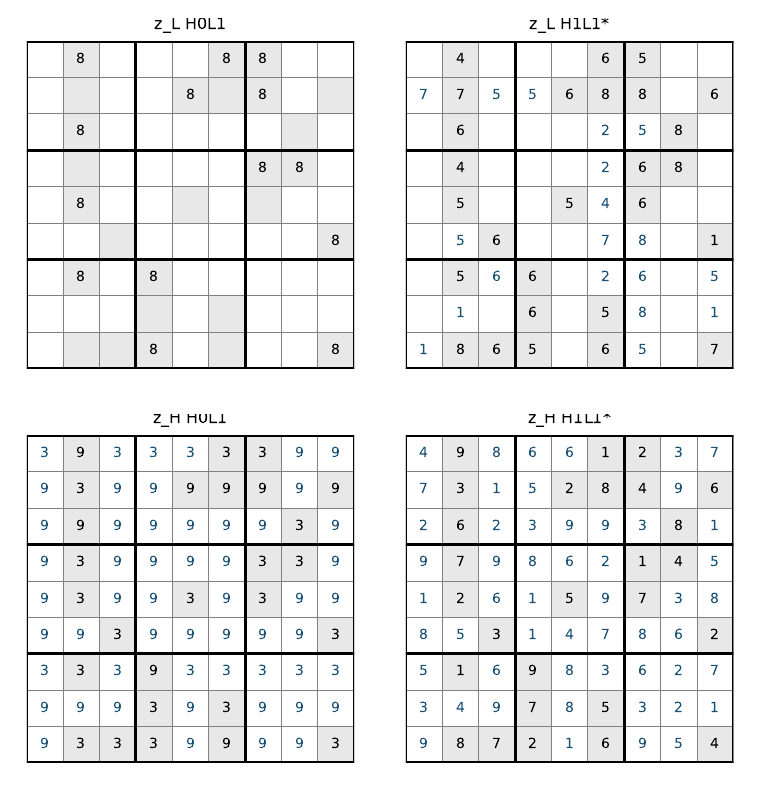}
  \captionof{figure}{Selected first-step cycle decodes for Sudoku HRM. The two
    columns show the H-cycle endpoints H0L1 and H1L1, with $z_L$ on top and
    $z_H$ below. The decoded grids are observational sanity checks rather than
    causal evidence.}
  \label{fig:app_cycle_decode_sudoku_hrm}
\end{center}

\begin{center}
  \textbf{TRM / Sudoku}\par\vspace{2pt}
  \includegraphics[width=0.80\textwidth]{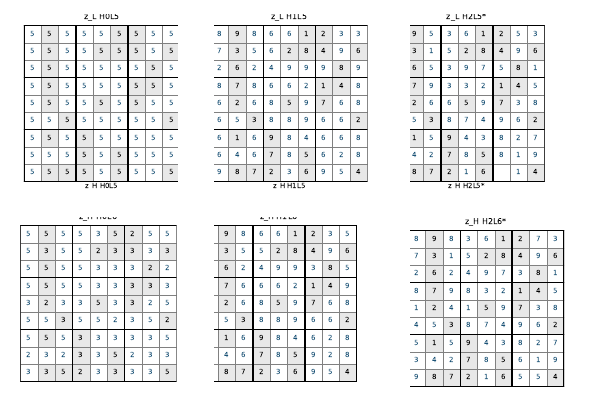}
  \captionof{figure}{Selected first-step cycle decodes for Sudoku TRM. The top
    row shows direct upstream $z_L$ cycles and the bottom row shows the
    corresponding $z_H$ updates.}
  \label{fig:app_cycle_decode_sudoku_trm}
\end{center}

\begin{center}
  \textbf{HRM / ARC-AGI}\par\vspace{2pt}
  \includegraphics[width=0.72\textwidth]{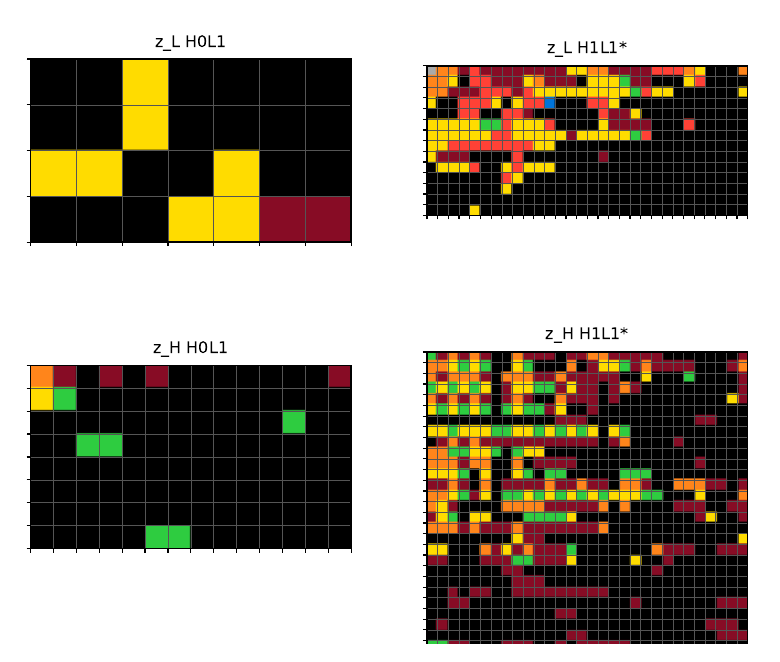}
  \captionof{figure}{Selected first-step cycle decodes for ARC-AGI HRM. The two
    columns show H-cycle endpoints, with $z_L$ on top and $z_H$ below.}
  \label{fig:app_cycle_decode_arc_hrm}
\end{center}

\begin{center}
  \textbf{TRM / ARC-AGI}\par\vspace{2pt}
  \includegraphics[width=0.80\textwidth]{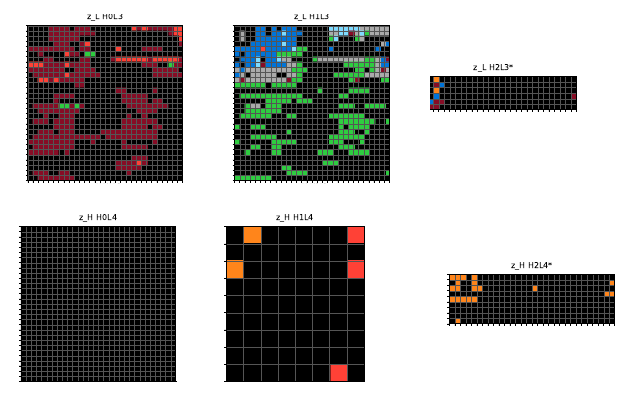}
  \captionof{figure}{Selected first-step cycle decodes for ARC-AGI TRM. As in
    Maze and Sudoku, these panels show what is decodable at each cycle, while
    the Jacobian analysis measures which positions influence later updates.}
  \label{fig:app_cycle_decode_arc_trm}
\end{center}

\section{Additional SAE locality results}
\label{app:sae}

\begin{center}
  \begin{minipage}{0.30\textwidth}
    \centering
    \includegraphics[width=\textwidth]{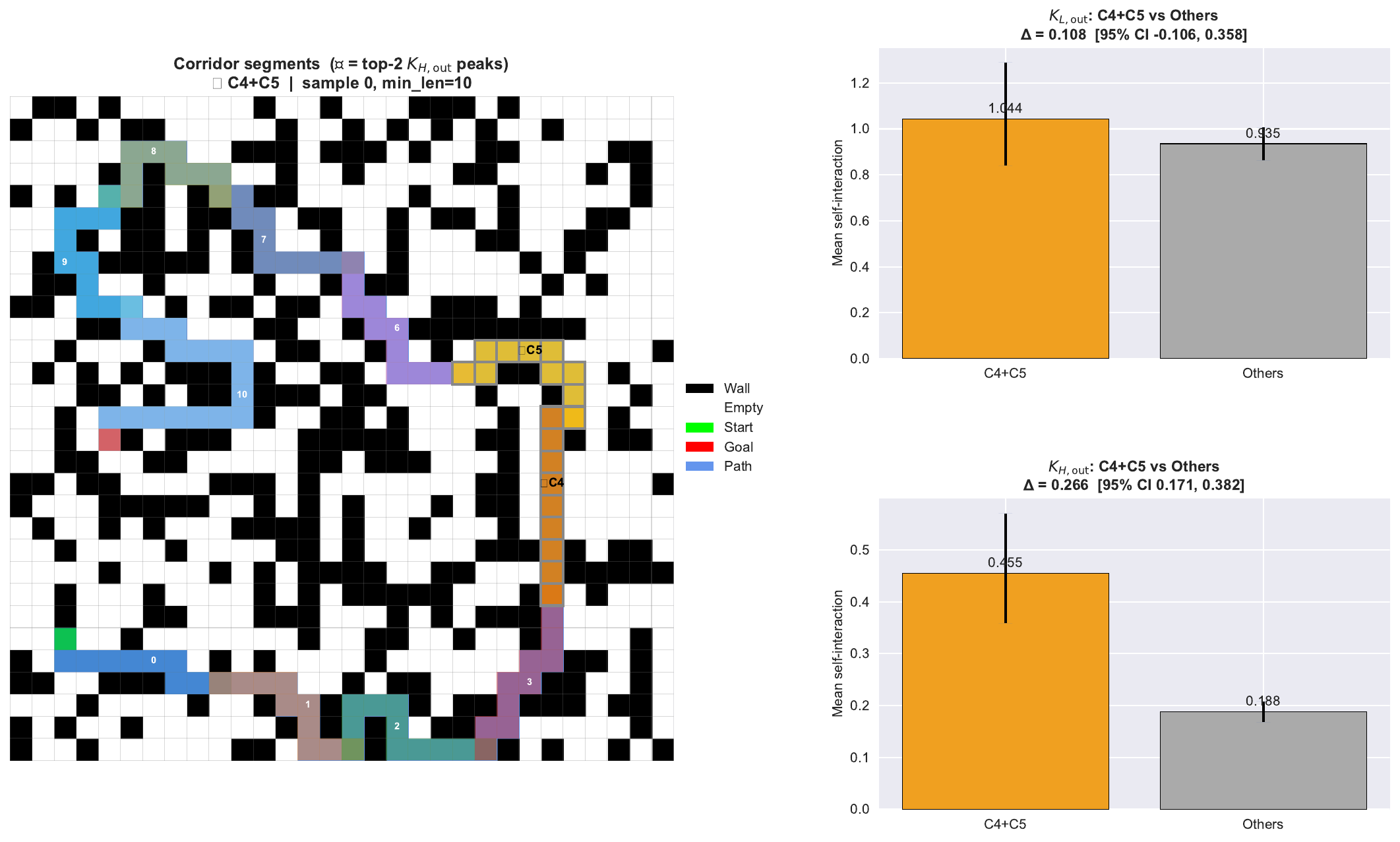}\\[-2pt]
    {\small (a) Maze HRM}
  \end{minipage}\hfill
  \begin{minipage}{0.30\textwidth}
    \centering
    \includegraphics[width=\textwidth]{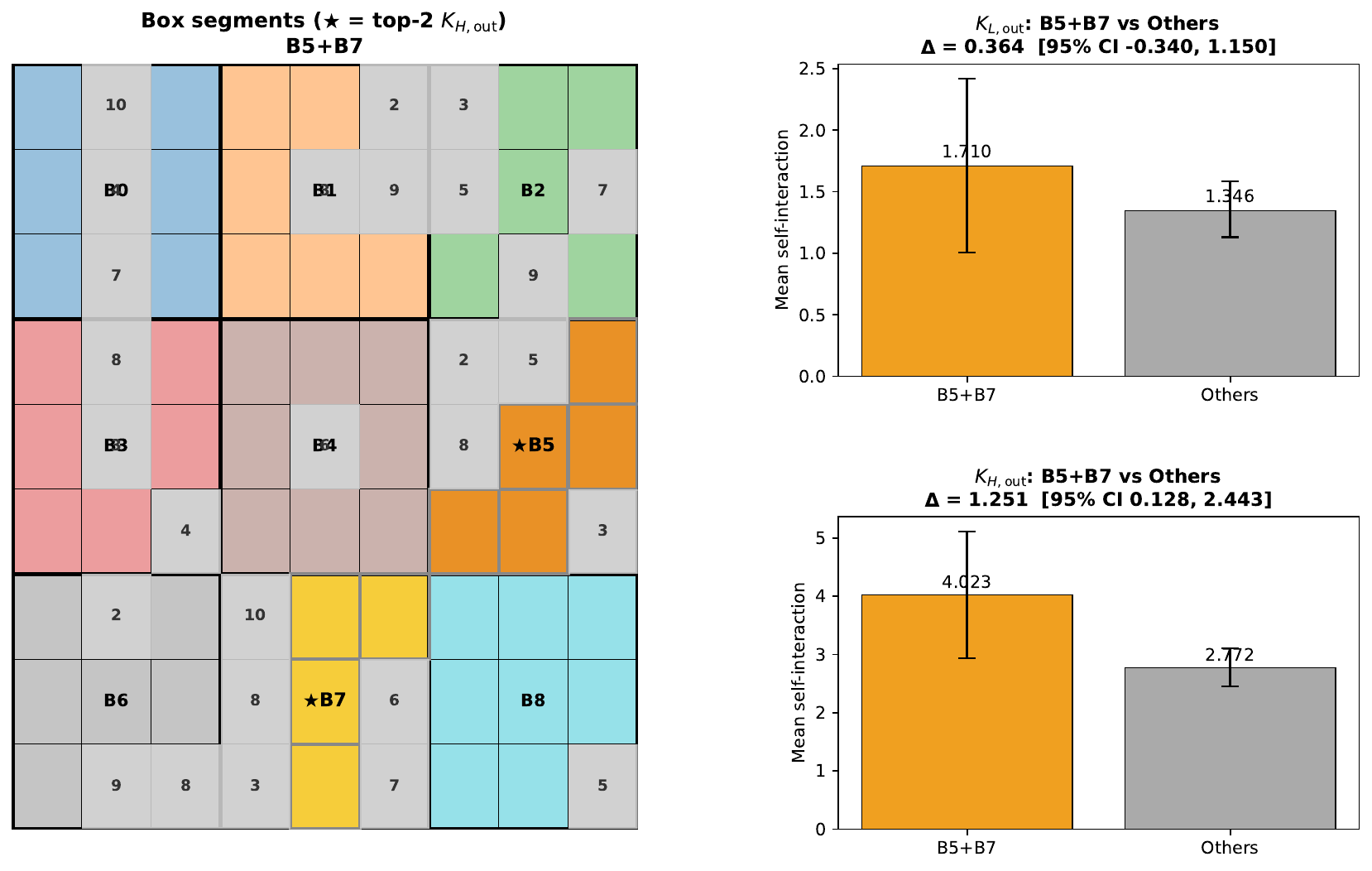}\\[-2pt]
    {\small (b) Sudoku HRM}
  \end{minipage}\hfill
  \begin{minipage}{0.30\textwidth}
    \centering
    \includegraphics[width=\textwidth]{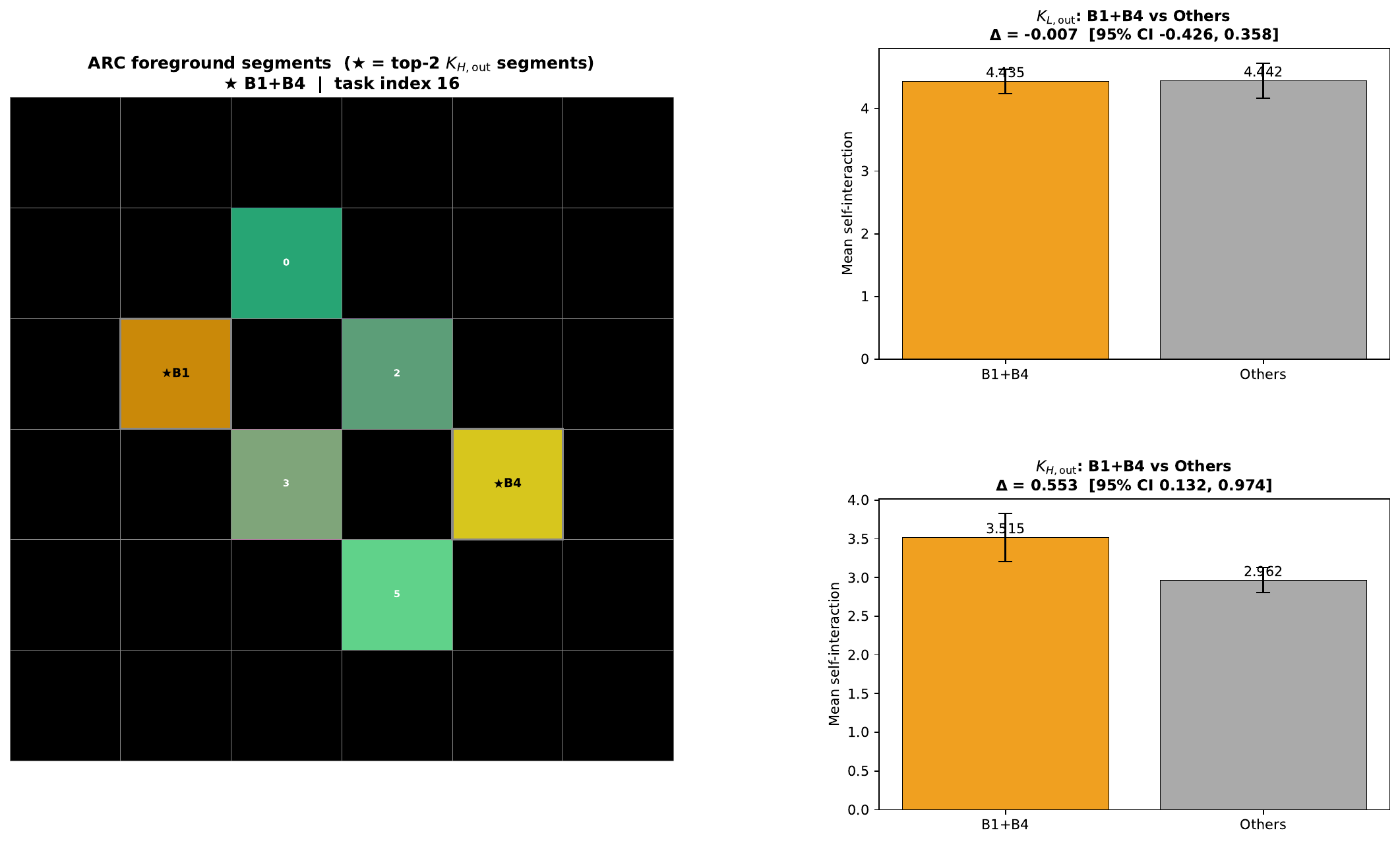}\\[-2pt]
    {\small (c) ARC-AGI HRM}
  \end{minipage}
  \\[8pt]
  \begin{minipage}{0.30\textwidth}
    \centering
    \includegraphics[width=\textwidth]{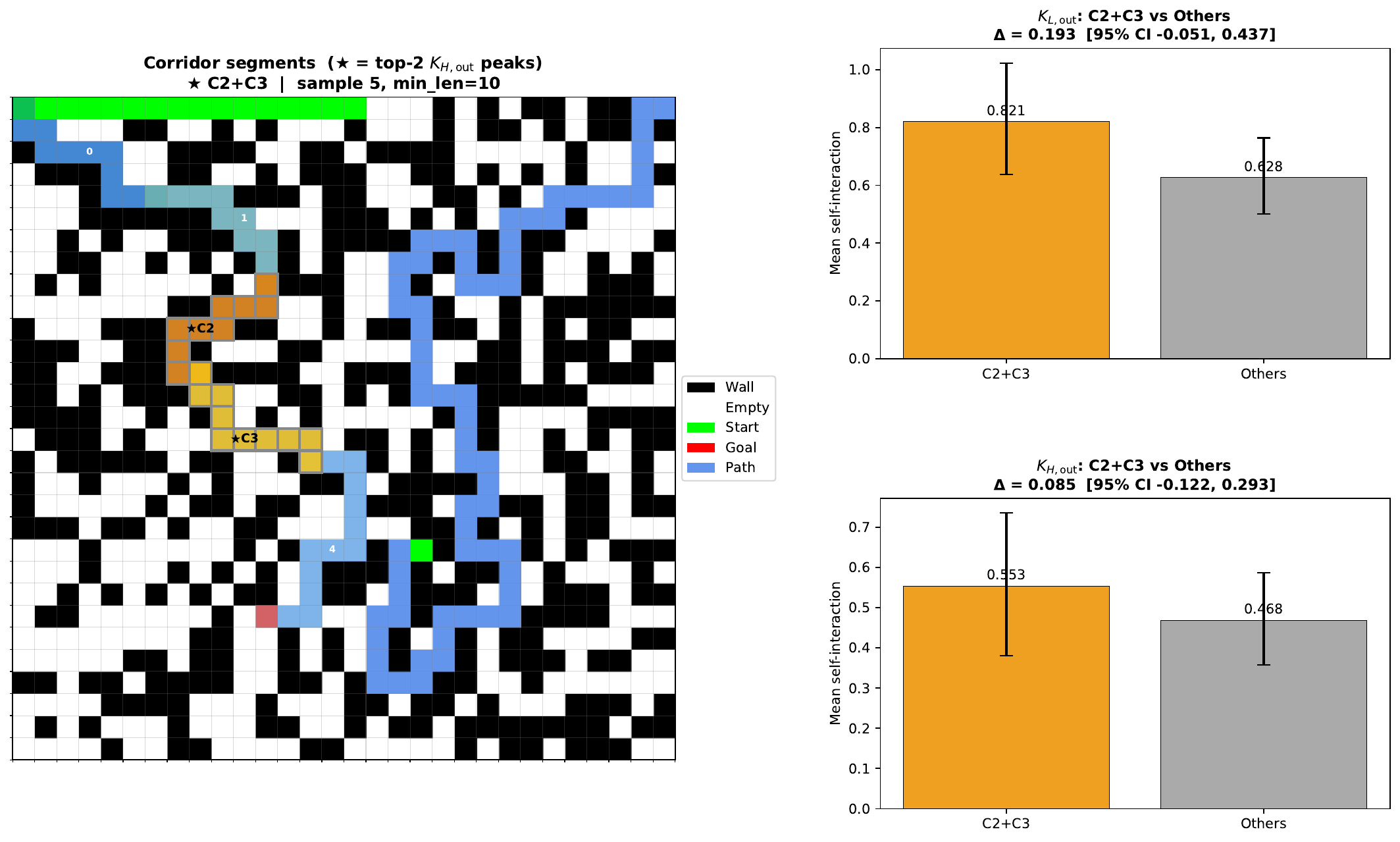}\\[-2pt]
    {\small (d) Maze TRM}
  \end{minipage}\hfill
  \begin{minipage}{0.30\textwidth}
    \centering
    \includegraphics[width=\textwidth]{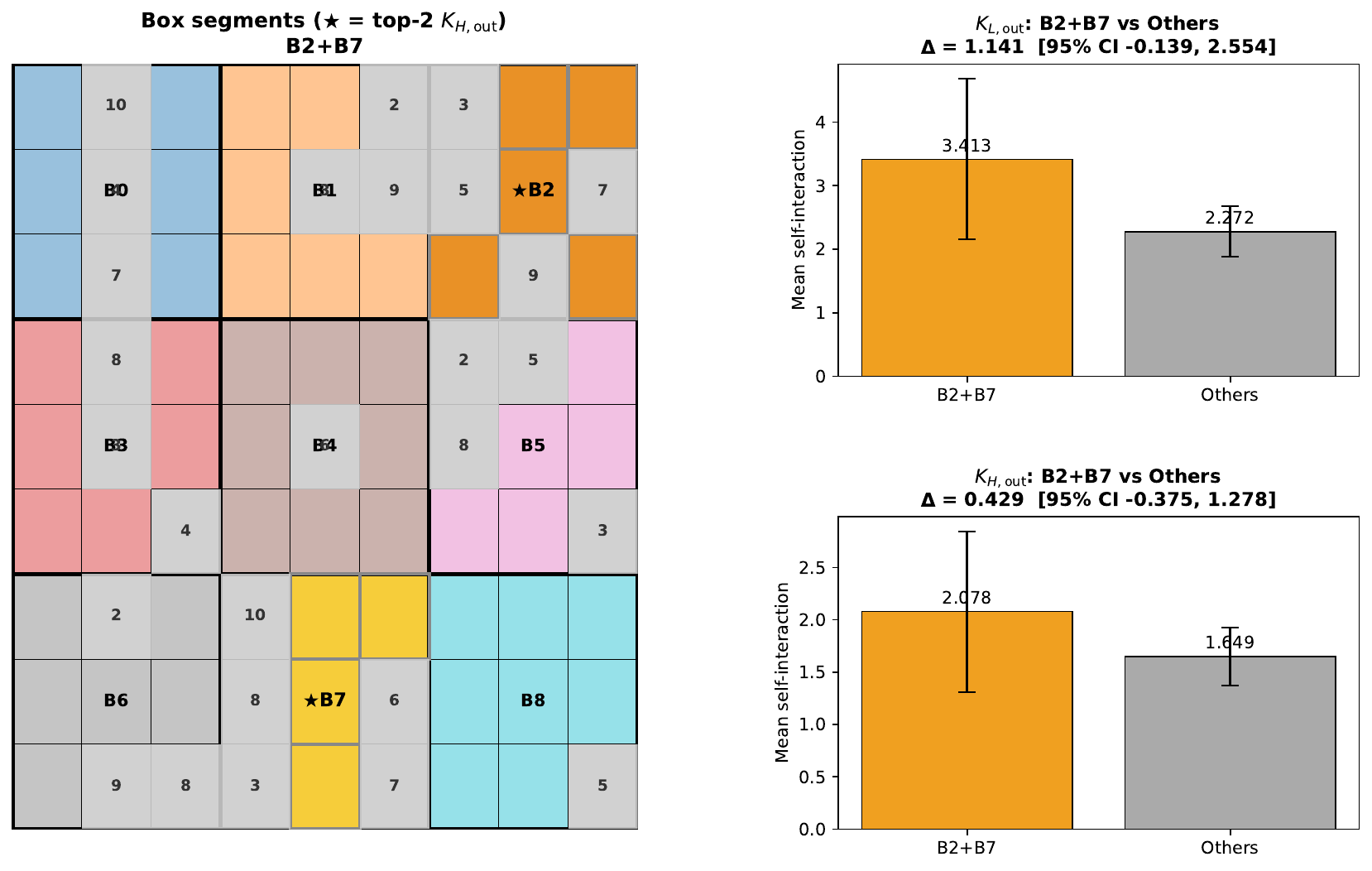}\\[-2pt]
    {\small (e) Sudoku TRM}
  \end{minipage}\hfill
  \begin{minipage}{0.30\textwidth}
    \centering
    \includegraphics[width=\textwidth]{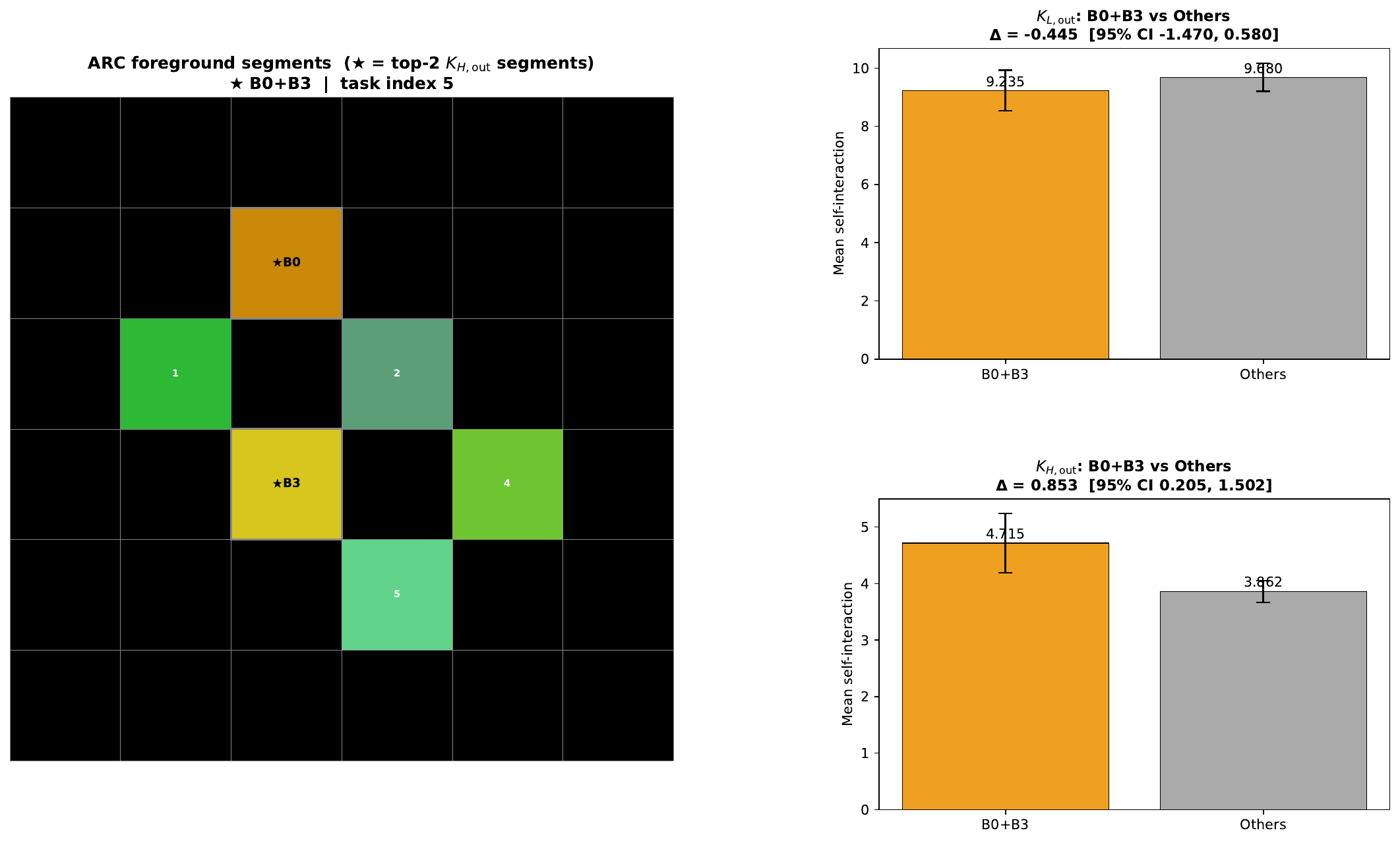}\\[-2pt]
    {\small (f) ARC-AGI TRM}
  \end{minipage}
  \captionof{figure}{Full top-segment qualitative analysis. The strongest human-readable positive cases are the HRM Maze and
    HRM Sudoku panels shown in the main text; the remaining panels show that
    many high-impact features are balanced or less cleanly aligned with the
    chosen segments.}
  \label{fig:app_top2_full}
\end{center}

\begin{center}
  \captionof{table}{Per-position correspondence between SAE feature locality and
    Jacobian locality. Correlations are computed from the cross-dataset summary
    files. Correlation is task- and architecture-dependent,
    reinforcing that SAE segment locality and Jacobian cell-to-cell locality are
    complementary rather than interchangeable measurements.}
  \label{tab:sae_jac_corr}
  \small
  \begin{tabular}{lccc}
    \toprule
    Configuration & $n$ & Pearson $r$ & Spearman $\rho$ \\
    \midrule
    Maze HRM & 10966 & .008 [n/a] & .009 [n/a] \\
    Maze TRM & 114 & .016 [-.126, .207] & .118 [-.073, .310] \\
    Sudoku HRM & 8100 & .478 [.459, .496] & .393 [.373, .412] \\
    Sudoku TRM & 8100 & .026 [.005, .047] & .021 [-.000, .043] \\
    ARC-AGI HRM & 2652 & .152 [.111, .193] & .140 [.102, .175] \\
    ARC-AGI TRM & 1768 & .510 [.471, .546] & .501 [.461, .538] \\
    \bottomrule
  \end{tabular}
\end{center}

\section{Additional within-cycle Jacobian and intervention results}
\label{app:jacobian}

The finite-noise patching summary in \Cref{tab:patching_summary} is the primary
within-cycle causal evidence in the main text. The Jacobian results below are
structural diagnostics: they summarize linearized cell-to-cell and segment-level
topology, and help explain why the finite-intervention patterns arise, but are
not treated as large-intervention causal effects.

For a Jacobian diagnostic, we write
$K^{b\leftarrow a}[u,v]=\|\partial z_b[u]/\partial z_a[v]\|_F$ for the local
linear influence from source site $v$ at level $a$ to target site $u$ at level
$b$. The shorthand $\KL$ and $\KH$ in the appendix figures denotes the L-output
and H-output kernels, respectively; it does not assume that $\KL$ is local or
that $\KH$ is global. Cell locality is measured after the fact by the
row-normalized diagonal concentration
$\bar{\ell}(K)=P^{-1}\sum_u K[u,u]/\sum_v K[u,v]$. For segment diagnostics, a
partition $\mathcal{S}=\{S_m\}$ induces
$K^{\mathrm{seg}}[m,n]=\mathrm{mean}_{u\in S_m,v\in S_n}K[u,v]$. When a
segment-granularity score is reported in supplementary outputs, we use
\begin{equation}
  g(K)=\frac{r_{\mathrm{seg}}}{1+\bar{\ell}(K)},
  \label{eq:granularity}
\end{equation}
where $r_{\mathrm{seg}}$ is the mean same-segment entry divided by the mean
cross-segment entry. The normalization keeps this descriptive segment score
from simply duplicating cell-diagonal concentration.

\paragraph{Within-cycle structural locality.}
\Cref{tab:app_jacobian_summary} restores the within-cycle Jacobian summary used
as a structural check. In HRM, the L-output kernel is more same-position local
than the H-output kernel in Maze and Sudoku, while ARC-AGI collapses the H/L gap
under object-local geometry. TRM changes this ordering: Maze and ARC-AGI have
larger H-output locality, while Sudoku is nearly tied. These linearized kernels
are not the main finite-causal evidence, but they show that the task geometry is
also visible in local differential topology.

\begin{center}
  \centering
  \captionof{table}{Within-cycle Jacobian same-position locality summary (95\%
  bootstrap CI). $\KL$ and $\KH$ are L-output and H-output Jacobian kernels,
  respectively; they are level-indexed structural diagnostics, not assumed local
  or global a priori. Sample sizes are $n=191$ for HRM Maze, $n=530$ for TRM
  Maze, and $n=100$ for the remaining model--task pairs.}
  \label{tab:app_jacobian_summary}
  \small
  \begin{tabular}{llccc}
    \toprule
    Architecture & Task & $\KL$ locality [95\% CI] & $\KH$ locality [95\% CI] & Ratio \\
    \midrule
    HRM & Maze    & .331 [.328, .333] & .225 [.220, .230] & 1.47$\times$ \\
    HRM & Sudoku  & .289 [.288, .289] & .205 [.204, .206] & 1.41$\times$ \\
    HRM & ARC-AGI & .541 [.492, .594] & .537 [.483, .597] & 1.01$\times$ \\
    \midrule
    TRM & Maze    & .576 [.575, .578] & .701 [.699, .704] & .82$\times$ \\
    TRM & Sudoku  & .372 [.371, .374] & .368 [.365, .370] & 1.01$\times$ \\
    TRM & ARC-AGI & .632 [.597, .673] & .681 [.647, .720] & .93$\times$ \\
    \bottomrule
  \end{tabular}
\end{center}

\paragraph{Segment granularity.}
\Cref{tab:app_segment_granularity} aggregates the same kernels into task
segments. Maze HRM remains L-over-H at both cell and corridor resolutions, while
Maze TRM is H-over-L; Sudoku is weaker but consistent with the nearly tied TRM
pattern; ARC-AGI shows that object-level segments can reverse the HRM cell-level
near tie. These results contextualize the main finite-noise patching results by
showing where the structural kernels place their segment mass.

\begin{center}
  \centering
  \captionof{table}{Segment-granularity summary with 95\% bootstrap CIs.
  Granularity is the segment self/cross ratio normalized by cell-level locality
  as in \Cref{eq:granularity}; higher values indicate influence concentrated
  within task segments but spread across cells inside those segments.}
  \label{tab:app_segment_granularity}
  \small
  \setlength{\tabcolsep}{4pt}
  \begin{tabular}{llccc}
    \toprule
    Architecture & Task & $g(\KL)$ [95\% CI] & $g(\KH)$ [95\% CI] & Direction \\
    \midrule
    HRM & Maze    & 7.35 [7.29, 7.41] & 4.35 [4.27, 4.43] & L $>$ H \\
    HRM & Sudoku  & 4.66 [4.62, 4.69] & 3.99 [3.96, 4.03] & L $>$ H \\
    HRM & ARC-AGI & 2.57 [2.30, 2.87] & 3.17 [2.82, 3.51] & H $>$ L \\
    \midrule
    TRM & Maze    & 12.08 [12.02, 12.14] & 15.60 [15.45, 15.74] & H $>$ L \\
    TRM & Sudoku  & 8.05 [7.96, 8.14] & 7.84 [7.74, 7.98] & near tie \\
    TRM & ARC-AGI & 4.53 [4.10, 4.99] & 6.29 [5.59, 7.07] & H $>$ L \\
    \bottomrule
  \end{tabular}
\end{center}

\begin{center}
  \centering
  \includegraphics[width=0.96\textwidth]{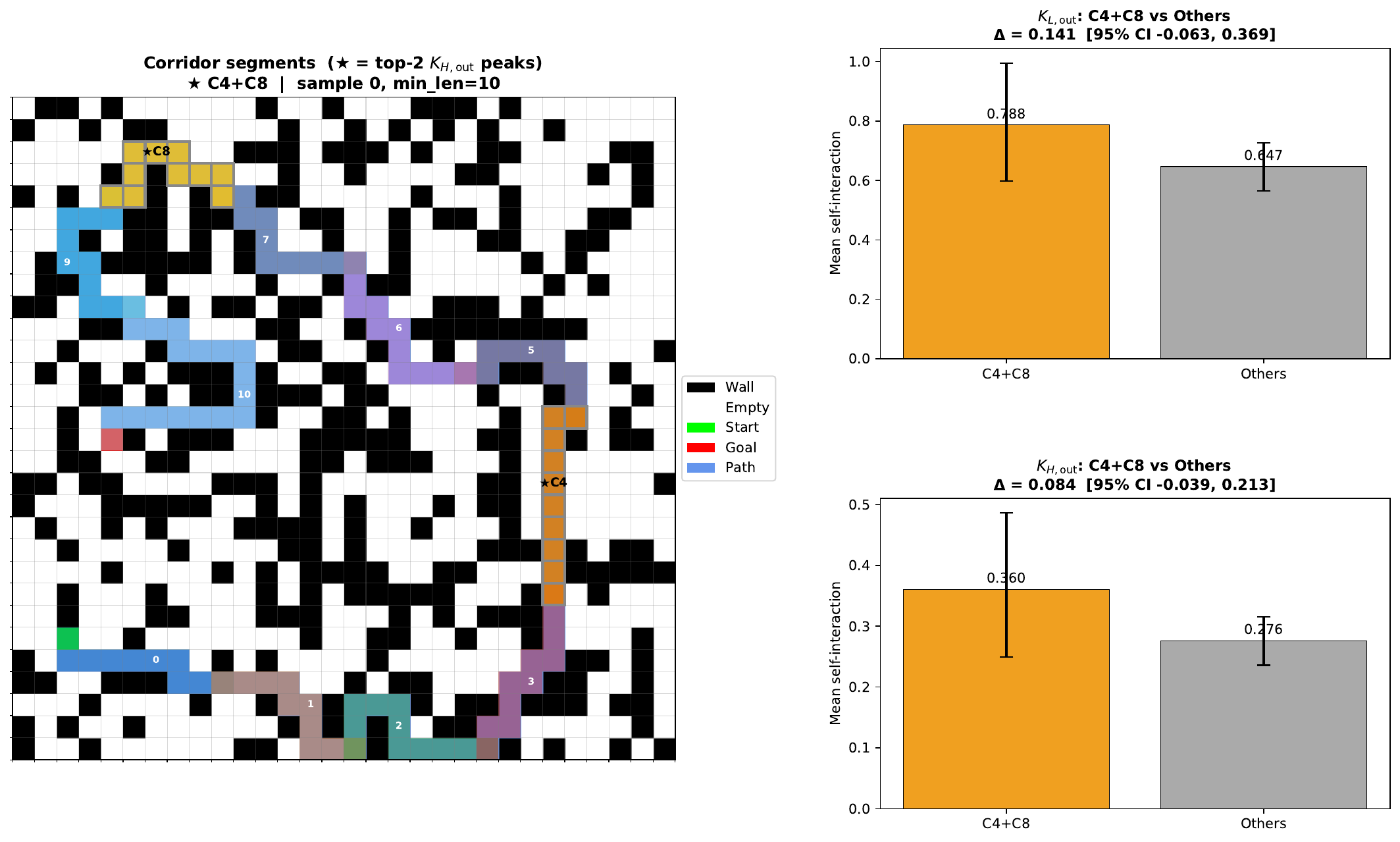}
  \captionof{figure}{Maze-TRM segment-level peak comparison. The left panel shows
    corridor segments on a representative maze, with the top two $\KH$ segment
    peaks highlighted; the right panels compare those peak segments against the
    remaining segments for $\KL$ and $\KH$. Maze-TRM localizes strong interaction structure to specific
    route segments, but confidence intervals overlap in this single-sample peak
    comparison, so aggregate statistics remain the primary evidence.}
  \label{fig:app_maze_trm_segment_peaks}
\end{center}

\begin{center}
  \centering
  \begin{minipage}{0.49\textwidth}
    \centering
    \includegraphics[width=\textwidth]{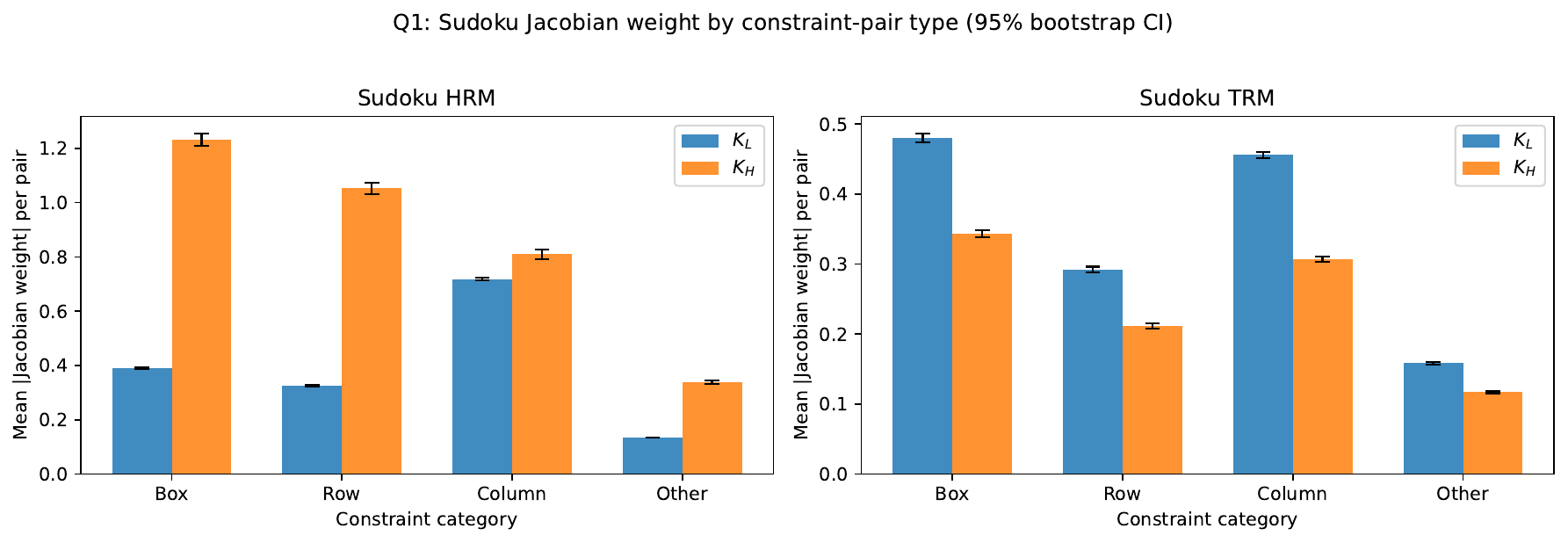}\par
    {\footnotesize\textbf{(a)} Sudoku constraint-pair breakdown.}
  \end{minipage}\hfill
  \begin{minipage}{0.47\textwidth}
    \centering
    \includegraphics[width=\textwidth]{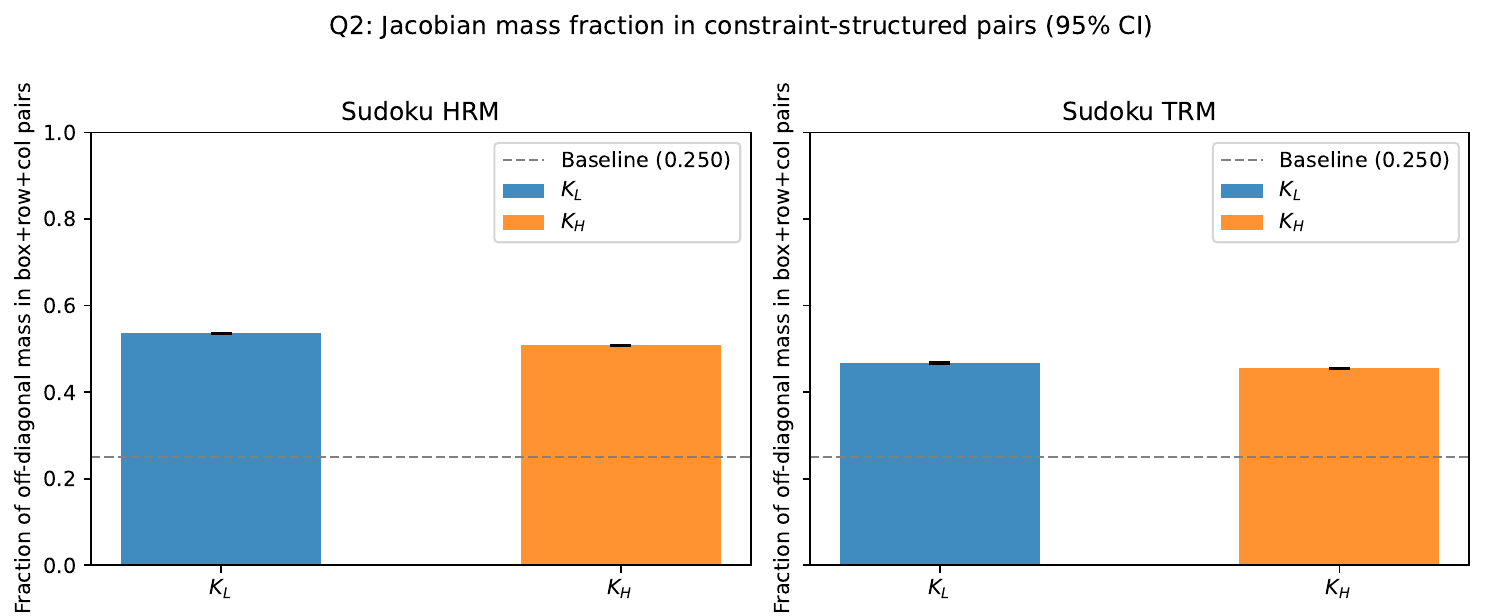}\par
    {\footnotesize\textbf{(b)} Sudoku structured-pair mass fraction.}
  \end{minipage}

  \vspace{0.9em}
  \includegraphics[width=0.70\textwidth]{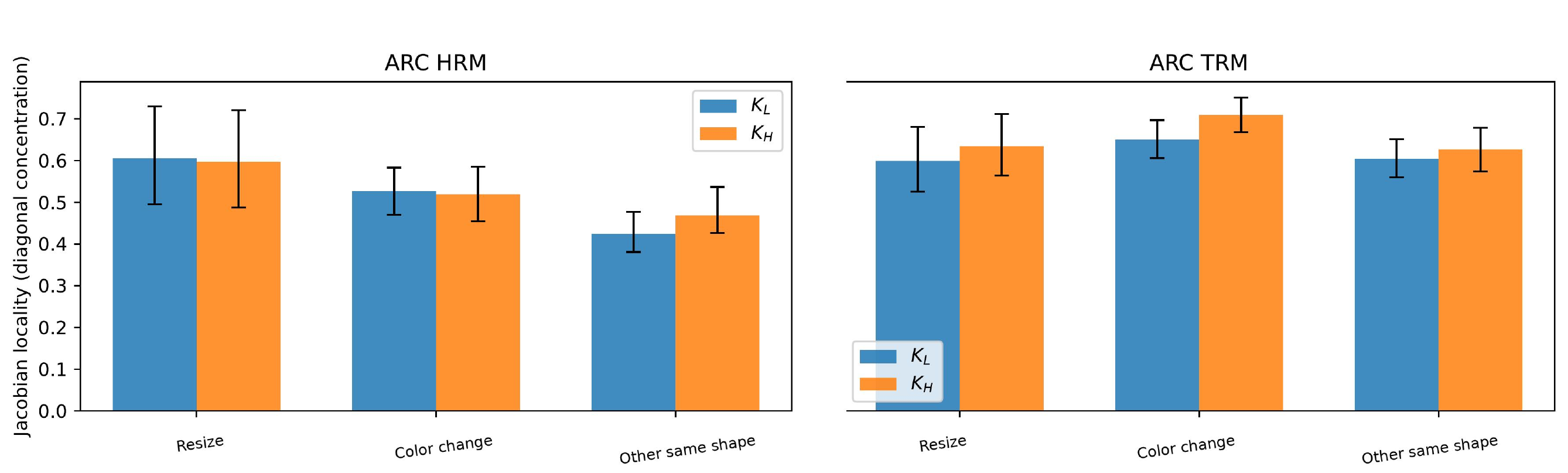}\par
  {\footnotesize\textbf{(c)} ARC-AGI locality by task type.}
  \captionof{figure}{Additional quantitative Jacobian segment analyses. Panel (a) breaks
    Sudoku Jacobian weight into box, row, column, and other cell-pair categories;
    panel (b) summarizes the off-diagonal mass in constraint-structured pairs;
    panel (c) stratifies ARC-AGI locality by task type.
    Segment-conditioned statistics expose structure that is not visible from the
    scalar cell-locality summary alone: HRM and TRM differ in how $\KL$ and $\KH$
    distribute mass across Sudoku constraint types, and ARC-AGI locality varies
    with task type.}
  \label{fig:app_jacobian_quant}
\end{center}

\begin{center}
  \centering
  \includegraphics[width=0.96\textwidth]{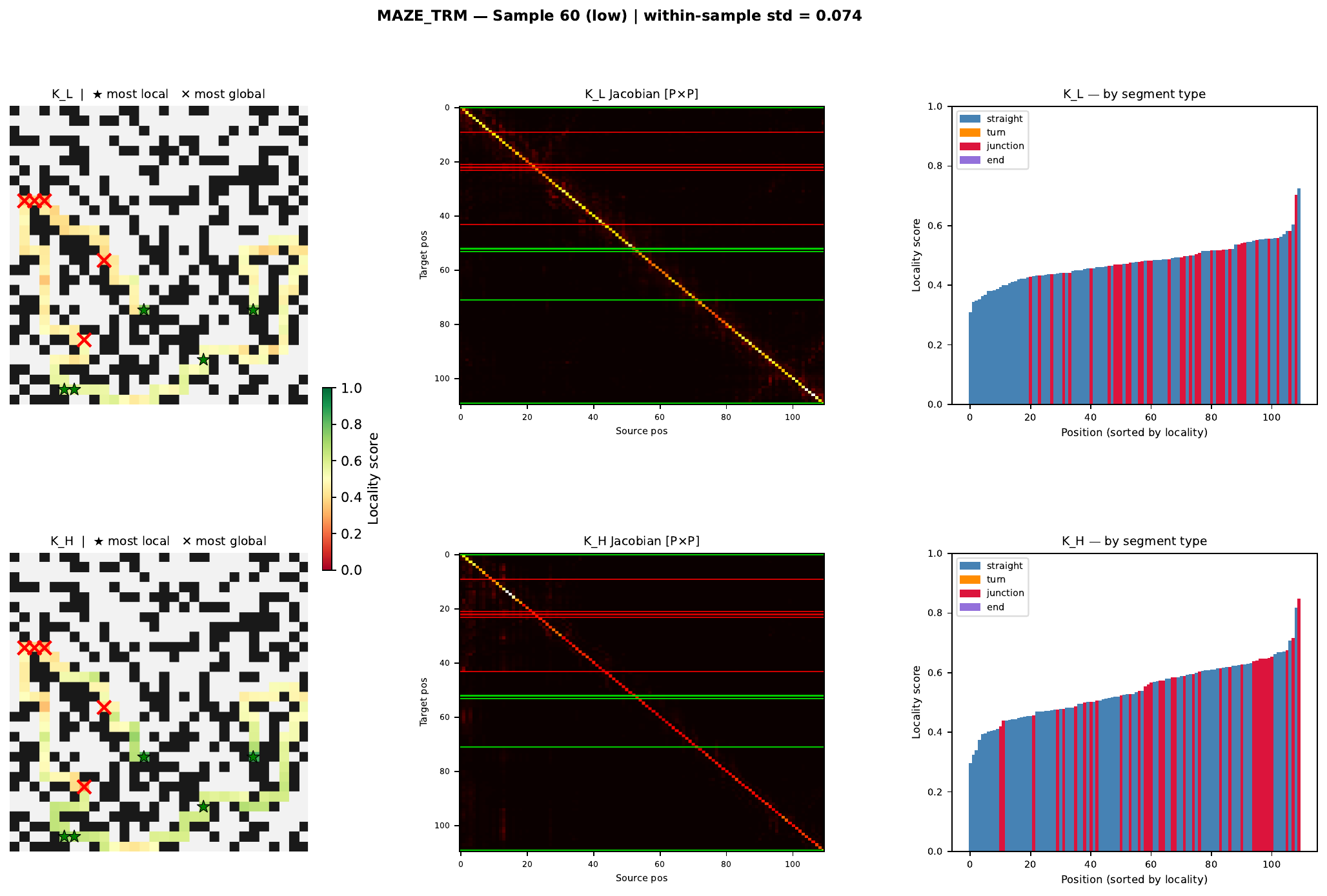}
  \captionof{figure}{Per-position Jacobian qualitative analysis for the selected
    high-variance Maze-TRM sample. Green marks more local positions and red marks
    more global positions. Locality is heterogeneous inside
    a single maze: some corridor stretches remain local while selected bottleneck
    or junction-adjacent positions act more globally.}
  \label{fig:app_jacobian_qual_maze}
\end{center}

\begin{center}
  \centering
  \includegraphics[width=0.96\textwidth]{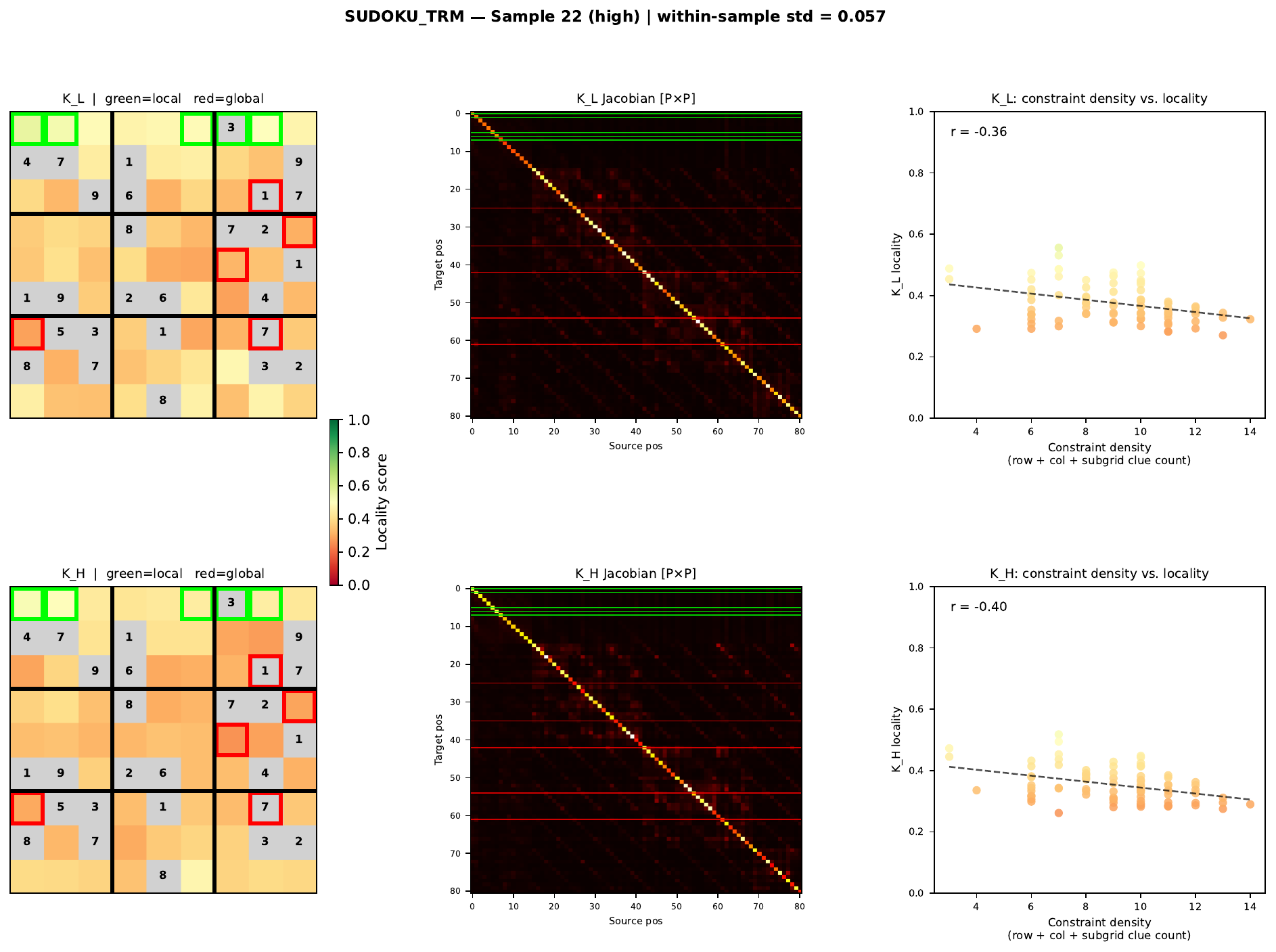}
  \captionof{figure}{Per-position Jacobian qualitative analysis for the selected
    high-variance Sudoku-TRM sample. The scatter plots compare constraint density
    against row-wise locality. In this sample, higher
    constraint density is associated with lower TRM locality for both $\KL$ and
    $\KH$, suggesting broader integration under heavier local constraints.}
  \label{fig:app_jacobian_qual_sudoku}
\end{center}

\begin{center}
  \centering
  \includegraphics[width=0.96\textwidth]{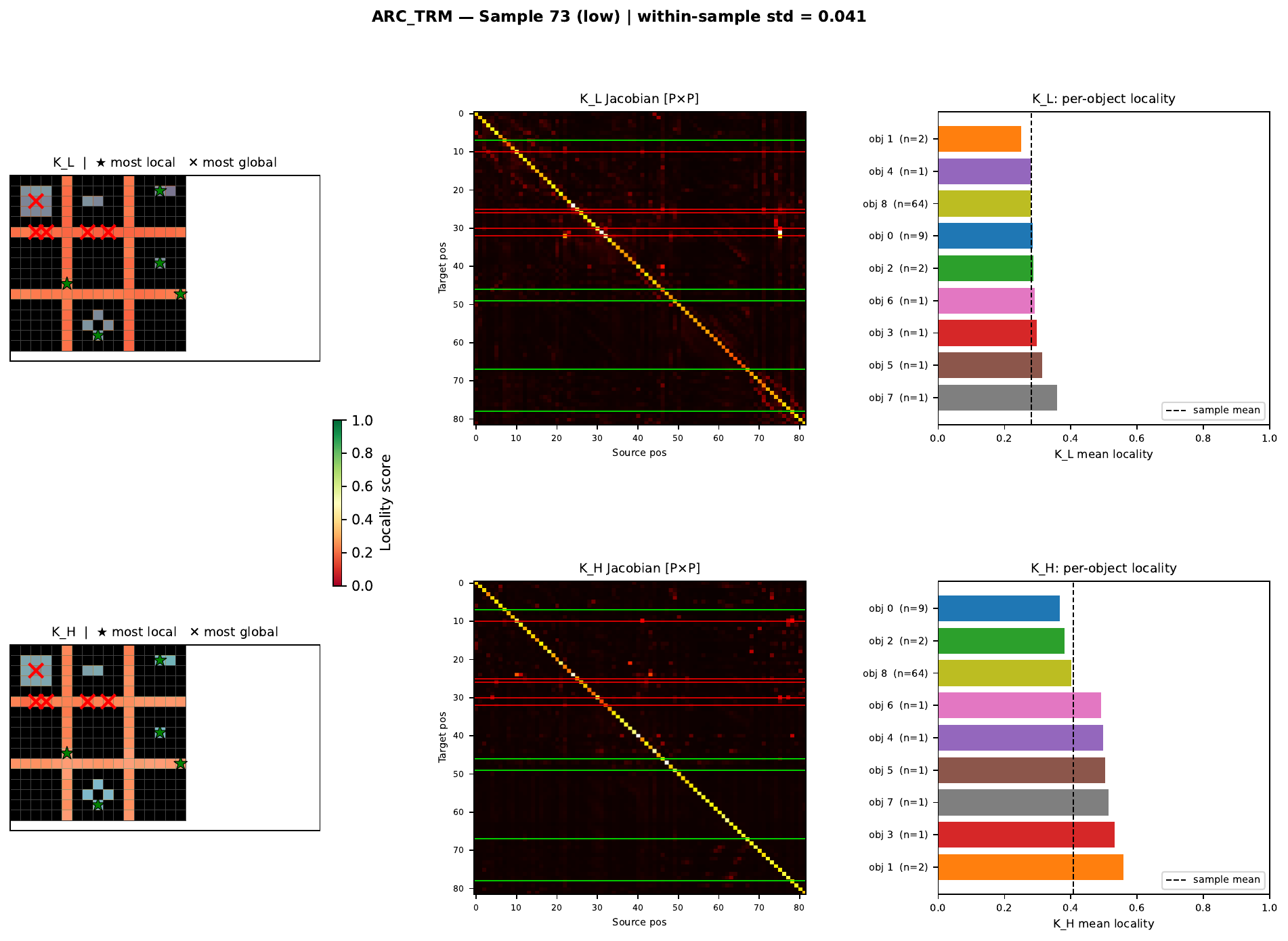}
  \captionof{figure}{Per-position Jacobian qualitative analysis for the selected
    high-variance ARC-TRM sample. Per-object bar charts summarize locality for
    connected foreground components. Object membership
    helps explain local/global variation, but the most global positions can lie on
    the task-defining object rather than only at isolated boundaries.}
  \label{fig:app_jacobian_qual_arc}
\end{center}

\section{Additional finite-noise activation patching results}
\label{app:patching}

\begin{center}
  \centering
  \includegraphics[width=0.95\textwidth]{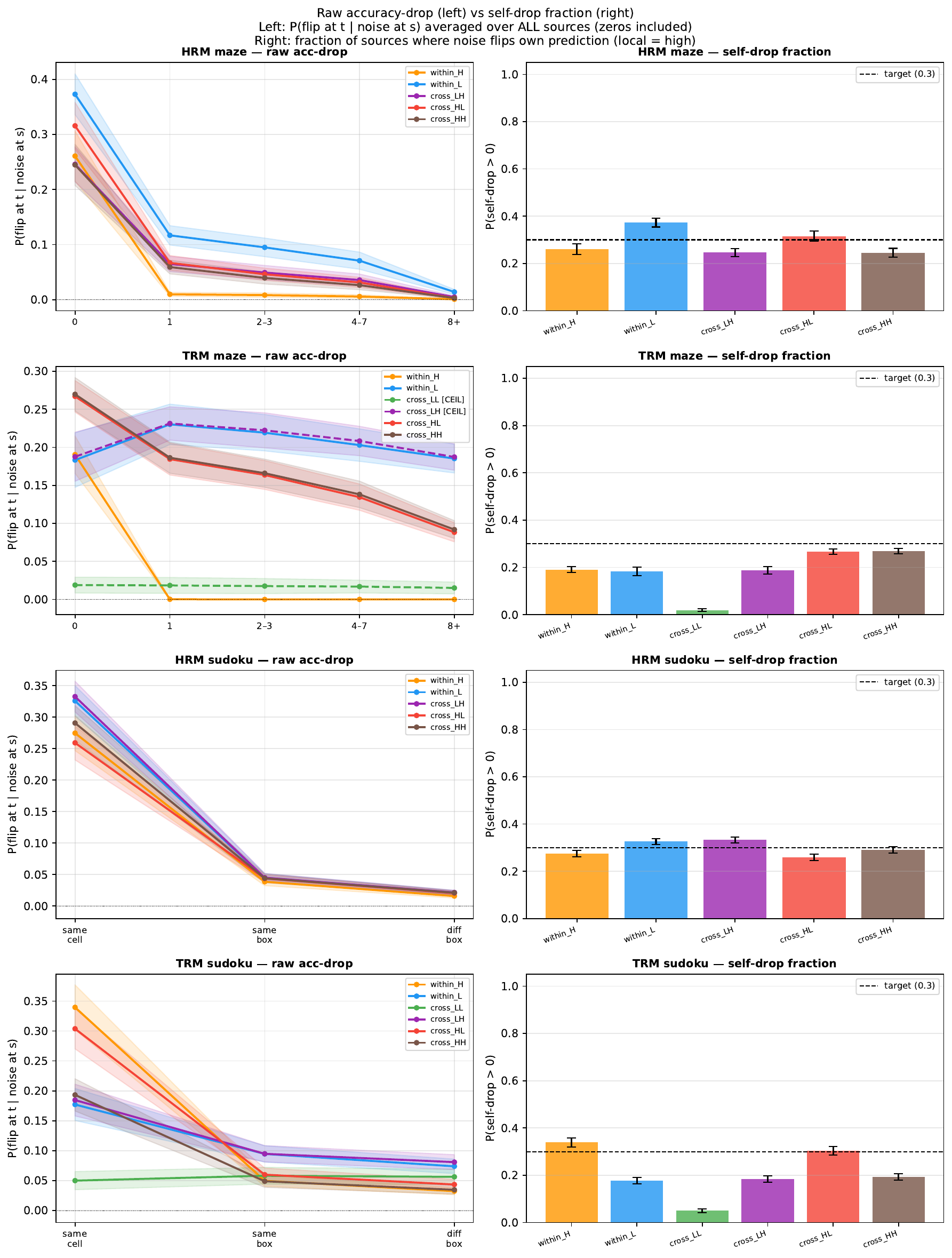}
  \captionof{figure}{Raw accuracy-drop curves and self-drop fractions for the
    finite-noise activation patching. Raw drops
    distinguish genuine spatial reach from globally diluted channels that can
    look misleading after diagonal normalization.}
  \label{fig:app_patching_raw}
\end{center}

\begin{center}
  \centering
  \includegraphics[width=0.98\textwidth]{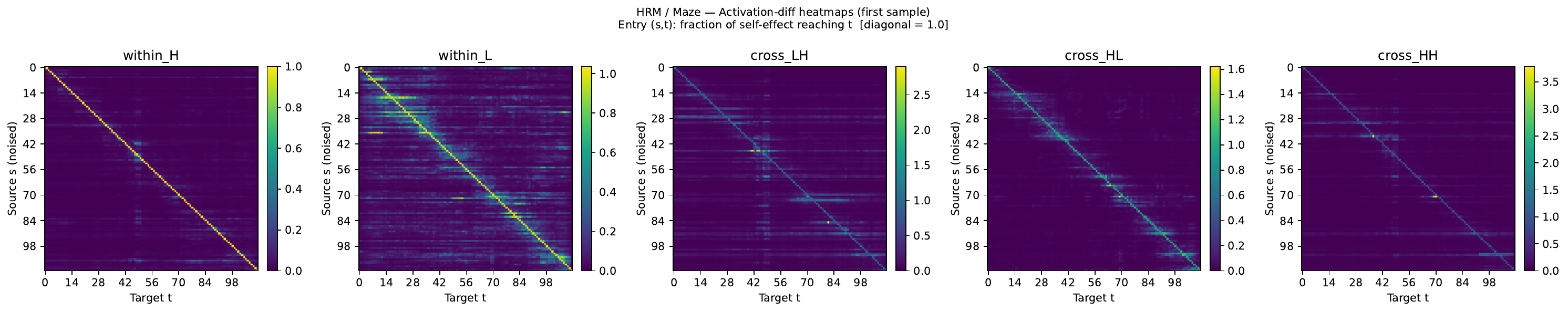}
  \captionof{figure}{Activation-difference heatmaps for HRM/Maze patching
    analogs. Rows are source positions and columns are target positions.
    Within-H is strongly diagonal, while cross-H channels
    carry broader off-diagonal effects across cycles.}
  \label{fig:app_patching_hrm_maze}
\end{center}

\begin{center}
  \centering
  \includegraphics[width=0.98\textwidth]{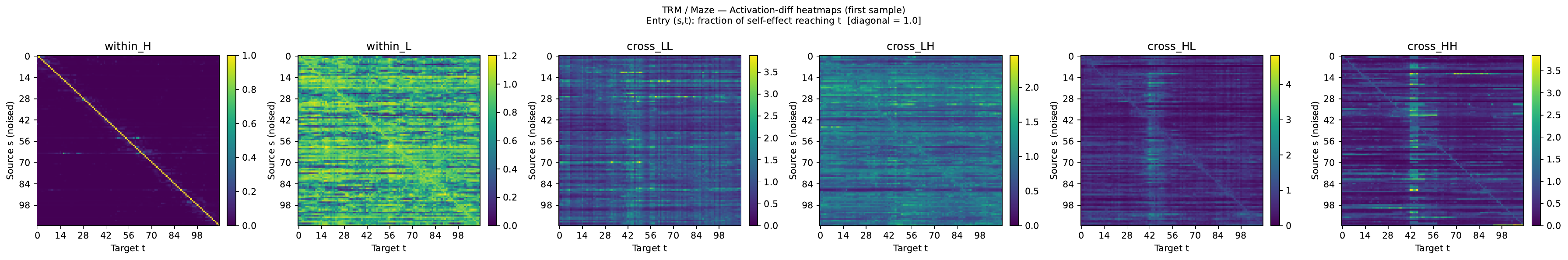}
  \captionof{figure}{Activation-difference heatmaps for TRM/Maze patching
    analogs. The within-L channel is broadly distributed,
    matching the main text's claim that weight sharing changes how local and
    global reach are expressed.}
  \label{fig:app_patching_trm_maze}
\end{center}

\begin{center}
  \centering
  \includegraphics[width=0.98\textwidth]{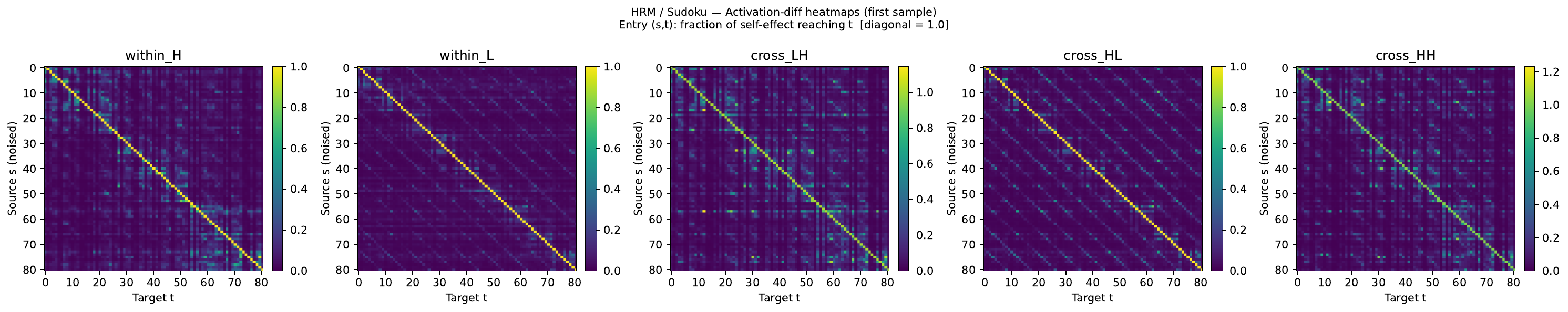}
  \captionof{figure}{Activation-difference heatmaps for HRM/Sudoku patching
    analogs. Sudoku shows structured constraint-related
    effects but weaker H/L separation than Maze.}
  \label{fig:app_patching_hrm_sudoku}
\end{center}

\begin{center}
  \centering
  \includegraphics[width=0.98\textwidth]{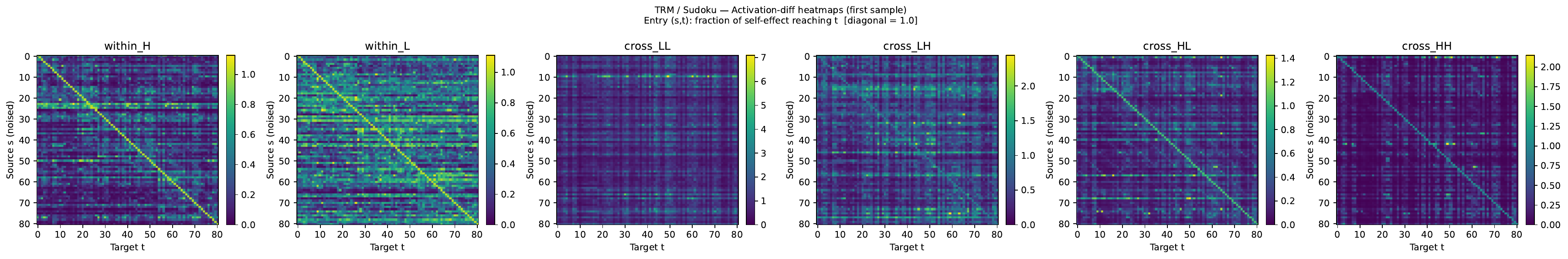}
  \captionof{figure}{Activation-difference heatmaps for TRM/Sudoku patching
    analogs. Several cross channels are broad or
    noise-ceiling cases, reinforcing the need for the reliability diagnostic in
    \Cref{fig:activation_patching_main}.}
  \label{fig:app_patching_trm_sudoku}
\end{center}

\section{Cross-dataset finite-noise patching locality}
\label{app:patching_cross_dataset}
\begin{center}
  \centering
  \includegraphicsmaybe[width=0.94\textwidth]{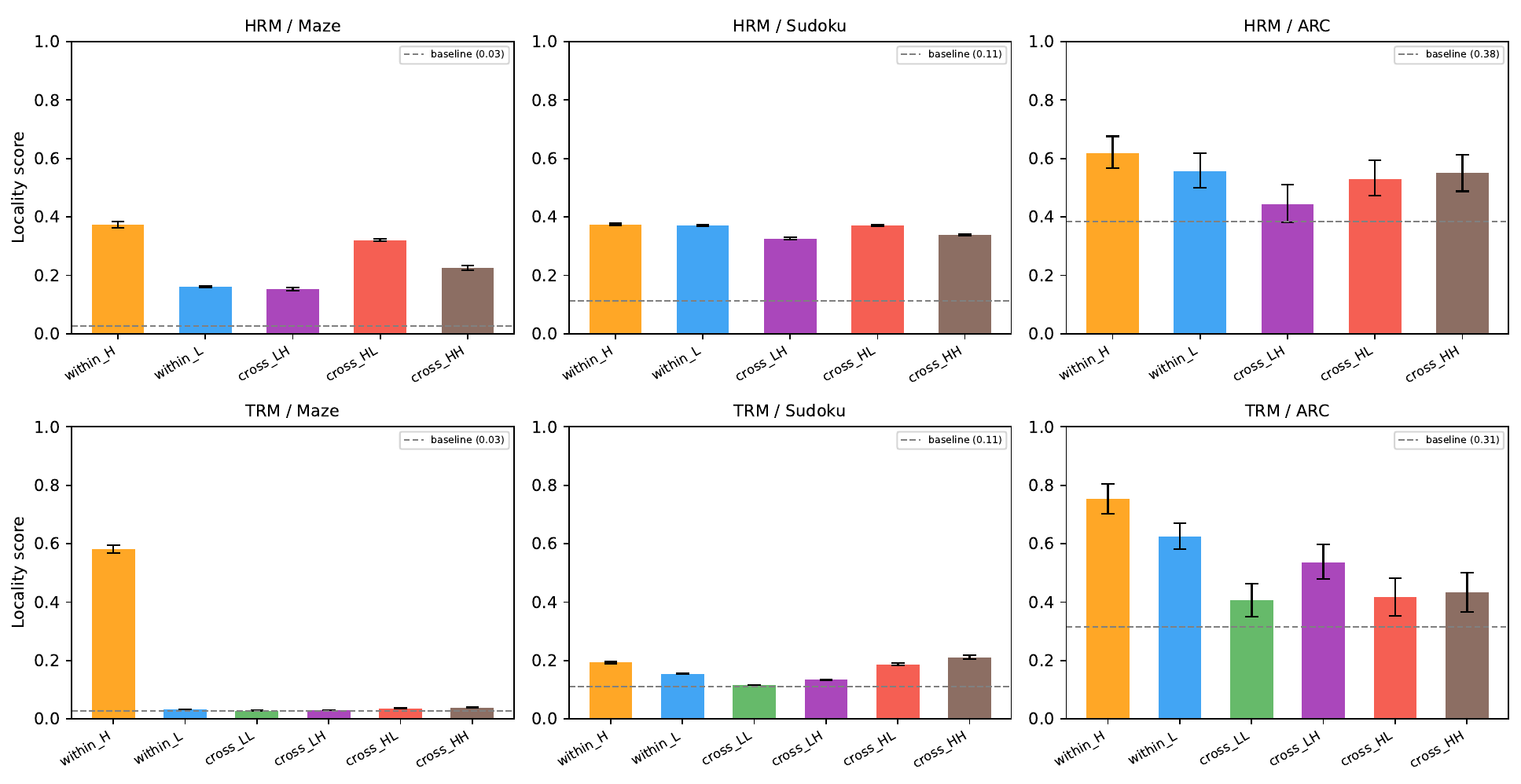}
  \captionof{figure}{Cross-dataset finite-noise patching locality from the experiment summarized numerically in \Cref{tab:patching_summary}. Within-H is at least as local as within-L under task-defined
  neighborhoods, while cross-cycle channels vary with architecture and task
  geometry.}
  \label{fig:app_p05_cross_dataset}
\end{center}

\section{Constraint-type breakdown: comparing Jacobian, patching, and SAE ablation}
\label{app:constraint_breakdown}
The three probe methods in our framework---structural Jacobian, finite-noise
activation patching, and SAE feature ablation---each make a different
commitment to causal strength, granularity, and scope. This section places them
side by side on the same question for Sudoku: \emph{does information flow respect
Sudoku constraint boundaries (box, row, column, or other)?}

For each cell pair $(i,j)$ in the $9\times9$ grid we assign a constraint label
based on which Sudoku rule links them: \emph{box} if they share the same
$3\times3$ subgrid (priority), \emph{row}/\emph{column} if they share a row/column
outside the same box, and \emph{other} otherwise.
The random-assignment baseline is box\,=\,0.100, row\,=\,0.075, col\,=\,0.075,
other\,=\,0.750 (exact counts from the $81\times80$ off-diagonal pair set).

\textbf{Jacobian} (structural): Per-cell pair, we compute the Frobenius-norm
Jacobian $K[i,j]=\|\partial z[i]/\partial z[j]\|_F$ at the critical cycle,
giving an (81,81) matrix per sample; fractions are sample-means weighted by
total mass.  The Jacobian figure covering HRM and TRM Sudoku is shown
in \Cref{fig:app_jacobian_quant}~(a).

\textbf{Activation patching} (causal): We use the averaged $(81\times81)$
heatmaps from the Jacobian-analog patching experiment; each entry is the mean impact on cell
$j$ when cell $i$ is source-patched.  Fractions are computed off-diagonal;
95\% CIs use pair-resampling bootstrap over all 6480 off-diagonal pairs.
Results are shown for the $K_H$- and $K_L$-analog channels (within-H and
within-L, respectively).

\textbf{SAE feature ablation} (causal, feature-resolved): For the top-30
H-level and L-level SAE features (by total ablation impact), we zero each
feature and record the change in output logits across all non-clue target
cells.  Impact is summed by constraint type; fractions average over features
and samples.

\Cref{fig:app_constraint_comparison} presents all three analyses side by side
for HRM (top row) and TRM (bottom row).  The structural
Jacobian $\KH$ shows strong constraint-type preferences---in particular,
$\KL$ concentrates substantially on \emph{column} pairs (${\approx}0.25$ vs.\
baseline $0.075$) in HRM---while both activation patching and SAE feature
ablation remain near the random baseline across all constraint types.  This
dissociation is informative: it shows that the cell-to-cell differential
topology aligns with Sudoku structure, but individual sparse features and
finite perturbations do not selectively activate along row/column lines.
The implication is that constraint-type selectivity in Sudoku is a collective
property of the full representational geometry rather than the province of
individually identifiable sparse features or activation-patching channels.

\begin{center}
  \centering
  \includegraphicsmaybe[width=\textwidth]{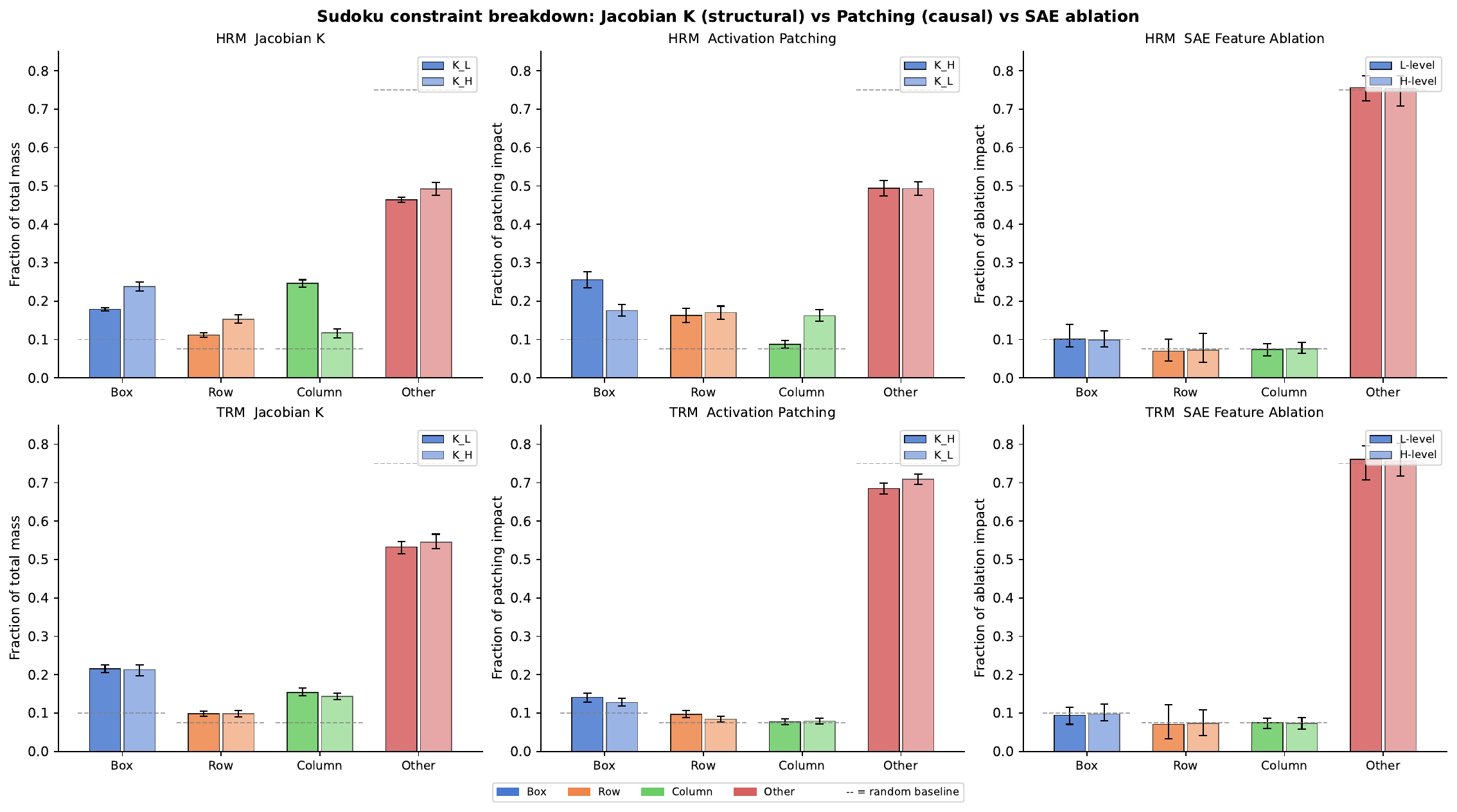}
  \captionof{figure}{Sudoku constraint-type breakdown for all three probe
    methods, HRM (top) and TRM (bottom).  \textbf{Col.\,1 (Jacobian~$K$):}
    structural Jacobian mass fraction by constraint type at the critical cycle;
    bars show $K_L$ (darker) and $K_H$ (lighter), error bars are 95\%
    bootstrap CI over samples.  \textbf{Col.\,2 (Activation patching):}
    off-diagonal fraction of the averaged impact heatmap by constraint type;
    $K_H$- and $K_L$-analog channels; error bars from pair-resampling
    bootstrap.  \textbf{Col.\,3 (SAE feature ablation):} mean constraint-type
    fraction across the top-30 H-level and L-level features; error bars from
    sample-resampling bootstrap.  Dashed lines mark the random pair-type
    baseline (box\,=\,0.10, row\,=\,col\,=\,0.075, other\,=\,0.75).
    The Jacobian shows clear constraint-type structure
    (especially column concentration in $\KL$), while patching and SAE
    ablation are near-baseline---indicating that Sudoku constraint-type
    selectivity is a collective geometric property rather than a signature of
    individual sparse features or finite perturbation channels.}
  \label{fig:app_constraint_comparison}
\end{center}

\section{Additional MTU3D locality results}
\label{app:mtu3d}

\begin{center}
  \centering
  \includegraphicsmaybe[width=0.88\textwidth]{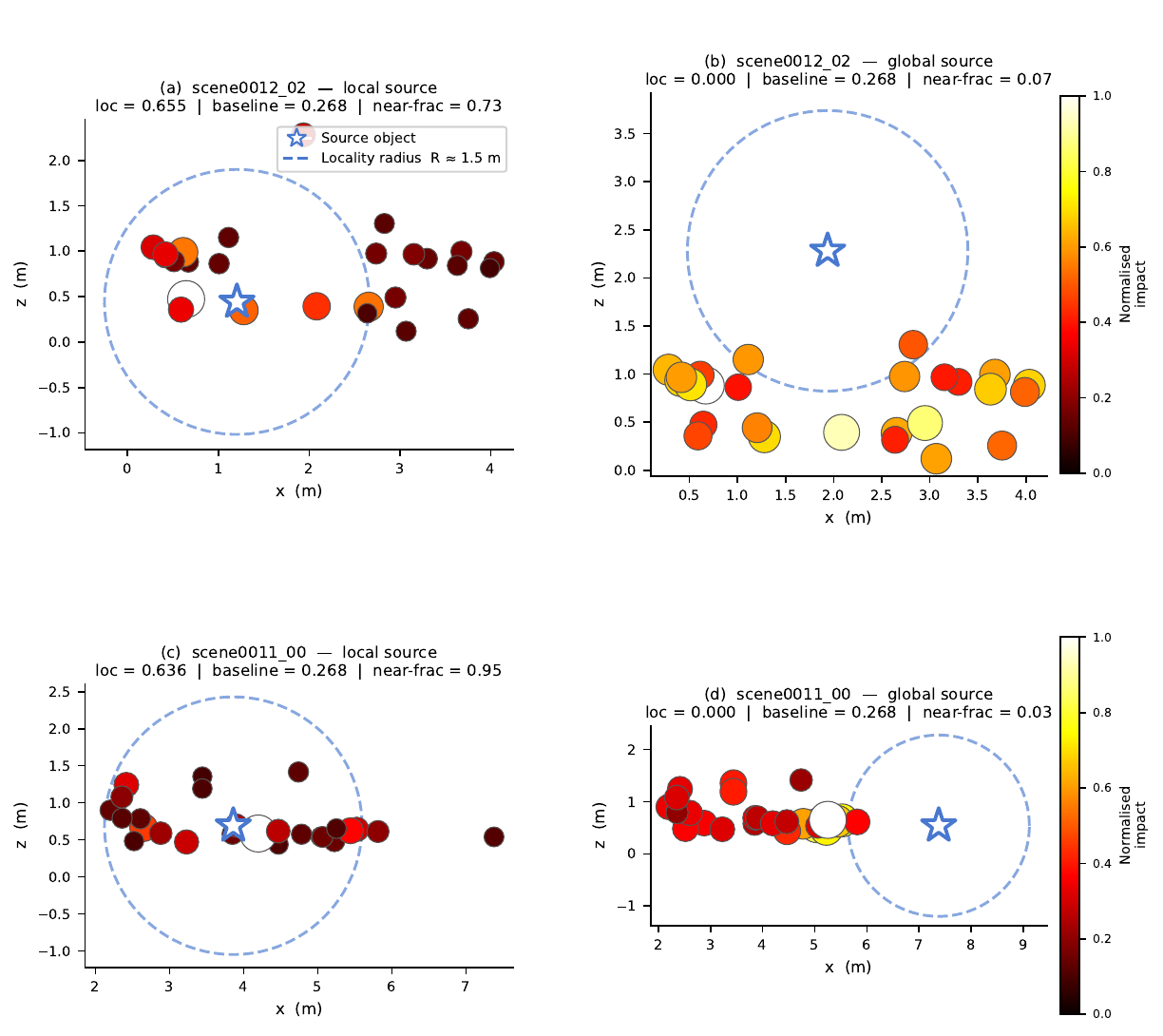}
  \captionof{figure}{Qualitative illustration of zero-ablation patching on two
  ScanNet scenes (top: \texttt{scene0012\_02}, bottom: \texttt{scene0011\_00};
  $N=27$ objects each).
  Each panel is a top-down $(x,z)$ projection of the scene.
  \textbf{Markers:} the white star ($\bigstar$) is the zero-ablated source object;
  all other objects are coloured and sized by the normalised activation-change norm
  $\|\Delta q\|_2$ (hot colourmap: bright/large = high impact, dark/small = low
  impact).
  The dashed circle shows the adaptive locality radius~$R$ (calibrated so that
  $\approx5$ nearest neighbours fall inside on average).
  \textbf{Near-frac} in the panel title is the fraction of total impact mass that
  falls within~$R$ of the source: it matches the per-object locality score used in
  the main-text Exp 1 bar chart.
  Local-source objects (left column; loc\,$\approx$0.64--0.66)
  concentrate impact within the radius (near-frac 0.73--0.95), while global-source
  objects (right column; loc\,$=$\,0.00) spread impact broadly across the scene
  (near-frac 0.03--0.07).
  Both extremes reflect real causal effects; the distinction is whether disruption
  is spatially structured or diffuse.
  This per-object variation underlies the aggregate locality score reported in the
  main text.}
  \label{fig:app_mtu3d_patching_spatial}
\end{center}

\begin{center}
  \centering
  \includegraphicsmaybe[width=0.94\textwidth]{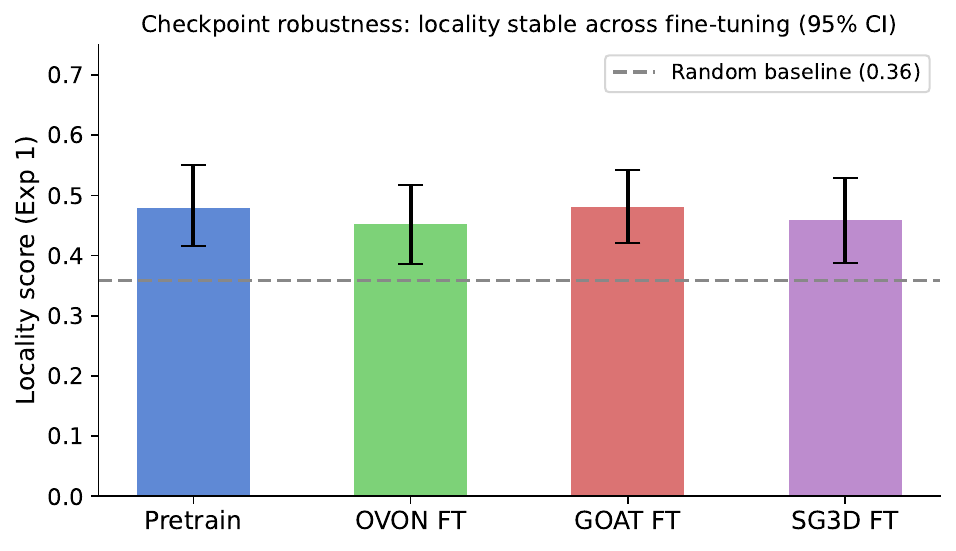}
  \captionof{figure}{MTU3D locality across checkpoints. Stage 1--2 input-patching locality is stable across checkpoints, whereas
  encoder-layer recovery remains near the random baseline.}
  \label{fig:app_mtu3d_checkpoint}
\end{center}

\begin{center}
  \centering
  \includegraphicsmaybe[width=0.94\textwidth]{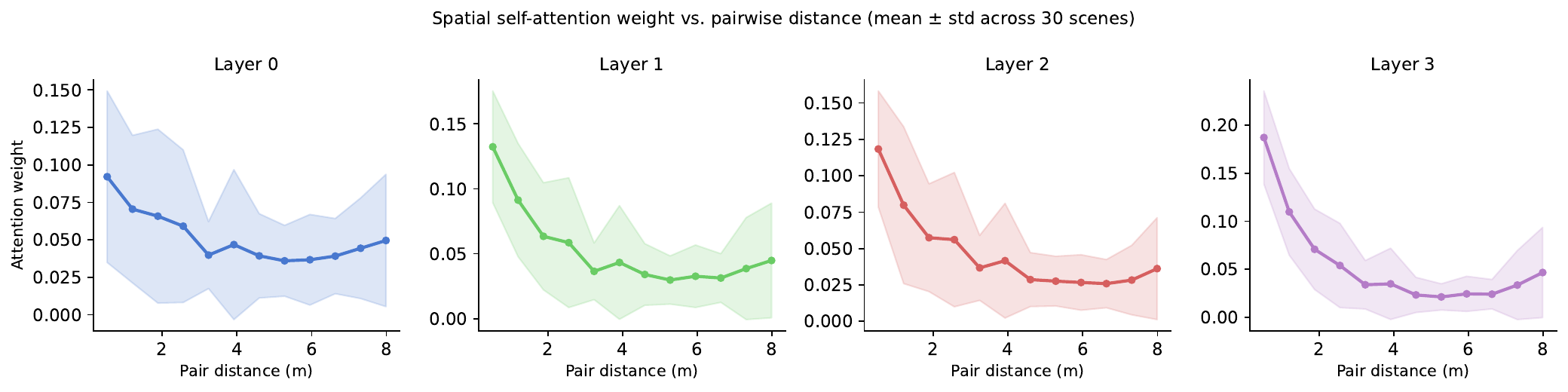}
  \captionof{figure}{MTU3D attention weight versus object-pair distance.
  Attention contains spatial distance bias, but the main
  text shows that such structural bias does not by itself imply finite causal
  recovery locality inside the unified encoder.}
  \label{fig:app_mtu3d_attention_distance}
\end{center}

\section{Additional cross-cycle Jacobian results}
\label{app:crosscycle}

For cross-cycle kernels, diagonal concentration means the same row-normalized
same-position mass $P^{-1}\sum_u K[u,u]/\sum_vK[u,v]$ computed between aligned
positions in adjacent recursive states. The heatmaps below visualize the full
row-normalized kernel; the tables report cross-cycle locality summaries and the
same-position concentration where H-to-H terms are available.

\begin{center}
  \centering
  \captionof{table}{Cross-cycle Jacobian locality by cycle edge (mean [95\% CI],
  $n=100$ per cell). For HRM, $H0L0\to H0L1$ is the within-first-H-step update,
  $H0L1\to H1L0$ is the cross-H-step boundary, and $H1L0\to H1L1$ is the
  within-second-H-step update. TRM rows use analogous late-within-H and boundary
  anchors, with exact edge labels shown because L-cycle counts differ across
  tasks.}
  \label{tab:app_crosscycle_summary}
  \scriptsize
  \setlength{\tabcolsep}{4pt}
  \begin{tabular}{llcc}
    \toprule
    Architecture & Task & Cycle edge & Cross-cycle locality [95\% CI] \\
    \midrule
    HRM & Maze   & $H0L0\to H0L1$ & .263 [.257, .270] \\
    HRM & Maze   & $H0L1\to H1L0$ & .261 [.256, .268] \\
    HRM & Maze   & $H1L0\to H1L1$ & .260 [.254, .266] \\
    \addlinespace[1pt]
    HRM & Sudoku & $H0L0\to H0L1$ & .264 [.263, .265] \\
    HRM & Sudoku & $H0L1\to H1L0$ & .268 [.267, .268] \\
    HRM & Sudoku & $H1L0\to H1L1$ & .243 [.242, .243] \\
    \addlinespace[1pt]
    HRM & ARC-AGI & $H0L0\to H0L1$ & .380 [.349, .414] \\
    HRM & ARC-AGI & $H0L1\to H1L0$ & .380 [.350, .413] \\
    HRM & ARC-AGI & $H1L0\to H1L1$ & .378 [.347, .411] \\
    \midrule
    TRM & Maze   & $H0L2\to H0L3$ & .501 [.499, .504] \\
    TRM & Maze   & $H0L3\to H1L0$ & .501 [.499, .503] \\
    TRM & Maze   & $H1L2\to H1L3$ & .499 [.497, .502] \\
    \addlinespace[1pt]
    TRM & Sudoku & $H0L4\to H0L5$ & .326 [.325, .328] \\
    TRM & Sudoku & $H0_{z_H}\to H1L0$ & .326 [.325, .328] \\
    TRM & Sudoku & $H1L4\to H1L5$ & .326 [.324, .328] \\
    \addlinespace[1pt]
    TRM & ARC-AGI & $H0L2\to H0L3$ & .513 [.485, .542] \\
    TRM & ARC-AGI & $H0L3\to H1L0$ & .514 [.486, .543] \\
    TRM & ARC-AGI & $H1L2\to H1L3$ & .513 [.485, .541] \\
    \bottomrule
  \end{tabular}
\end{center}

\begin{center}
  \centering
  \captionof{table}{Available $H{\to}H$ cross-cycle diagonal concentrations (mean [95\% CI], $n=100$ per cell). Values are
    listed as sequences over H-update pairs when present in the updated file.
    Maze and Sudoku TRM have no comparable $H{\to}H$ term under the analyzed
    L-cycle indexing, while ARC-AGI TRM has $z_H$ boundary entries.}
  \label{tab:app_crosscycle_hh}
  \scriptsize
  \setlength{\tabcolsep}{4pt}
  \begin{tabular}{lll}
    \toprule
    Architecture & Task & $H{\to}H$ diagonal concentration sequence \\
    \midrule
    HRM & Maze   & .900 [.899, .902] $\to$ .462 [.452, .474] \\
    HRM & Sudoku & .480 [.480, .480] $\to$ .218 [.216, .220] \\
    HRM & ARC-AGI & .736 [.698, .775] $\to$ .650 [.604, .696] $\to$ .517 [.461, .573] $\to$ .507 [.450, .563] \\
    \midrule
    TRM & Maze   & -- \\
    TRM & Sudoku & -- \\
    TRM & ARC-AGI & .754 [.724, .781] $\to$ .659 [.623, .693] $\to$ .602 [.560, .642] \\
    \bottomrule
  \end{tabular}
\end{center}

\begin{center}
  \centering
  \includegraphics[width=0.98\textwidth]{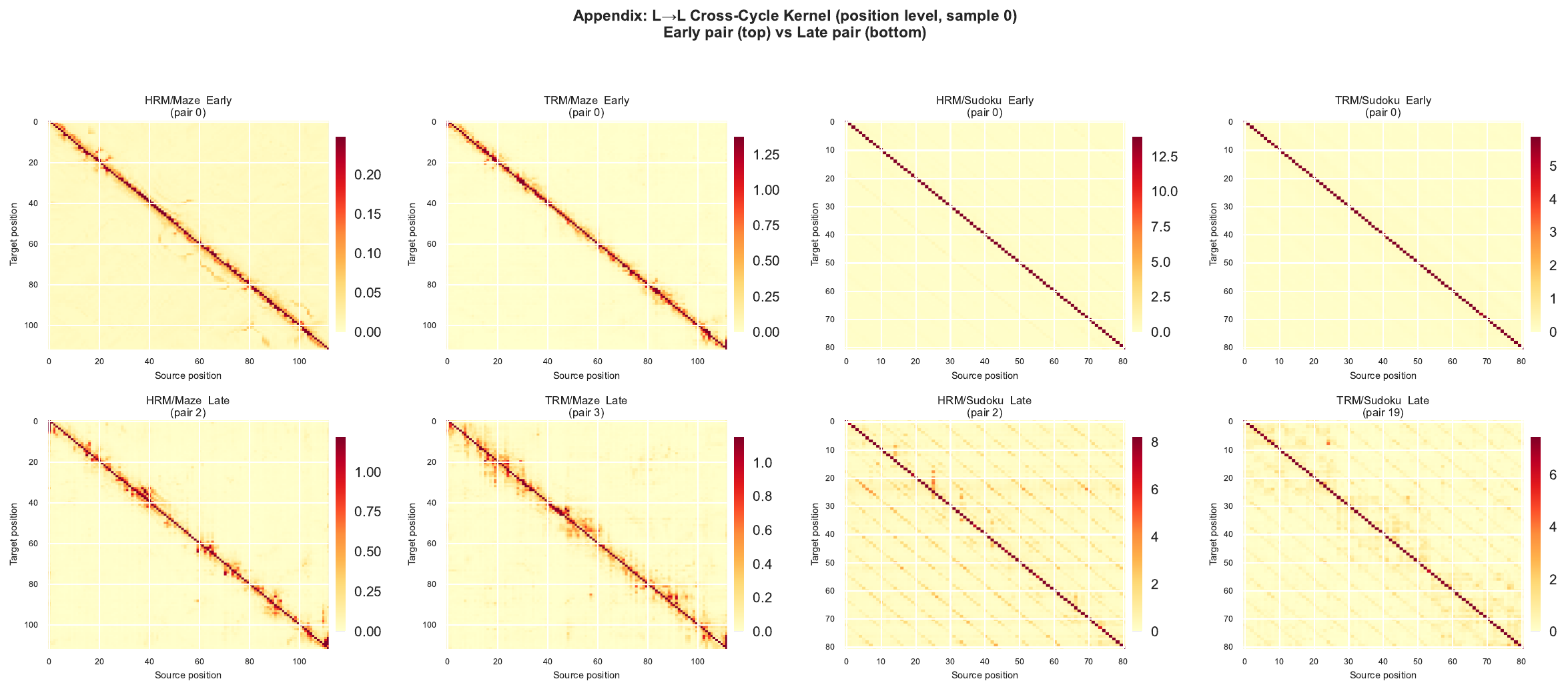}
  \captionof{figure}{Position-level L$\to$L cross-cycle Jacobian heatmaps for early and late
    cycle pairs. The heatmaps visualize the same transition
    captured by same-position diagonal concentration: early pairs are more self-focused and
    later pairs show broader spatial attribution.}
  \label{fig:app_cc_heatmaps}
\end{center}

\Cref{tab:app_crosscycle_hh,fig:app_cc_heatmaps} give additional quantitative
and spatial detail for the appendix structural cross-cycle analysis.

\end{document}